\documentclass[10pt,twocolumn,letterpaper]{article}

\usepackage[pagenumbers]{iccv} % To force page numbers, e.g. for an arXiv version

\usepackage[linesnumbered]{algorithm2e}
\usepackage{multirow}
\usepackage{graphics}

\definecolor{iccvblue}{rgb}{0.21,0.49,0.74}
\usepackage[pagebackref,breaklinks,colorlinks,allcolors=iccvblue]{hyperref}

\def\paperID{11511} % *** Enter the Paper ID here
\def\confName{ICCV}
\def\confYear{2025}

\title{DeSPITE: Exploring Contrastive \underline{\textbf{De}}ep \underline{\textbf{S}}keleton-\underline{\textbf{P}}ointcloud-\underline{\textbf{I}}MU-\underline{\textbf{T}}ext \underline{\textbf{E}}mbeddings for Advanced Point Cloud Human Activity Understanding}

\author{
Thomas Kreutz$^{1}$ \quad
Max Mühlhäuser$^{1}$ \quad
Alejandro Sanchez Guinea$^{2}$ \\
$^{1}$ Telekooperation Lab, Technical University Darmstadt \\
$^{2}$ NTT DATA, Luxembourg \\
{\tt\small \{kreutz@tk, max@informatik\}.tu-darmstadt.de, alejandro.guinea@global.ntt}
}

\begin{document}
\maketitle
\begin{abstract}
    % Human activities have a natural correspondence when recorded through visual and inertial sensors, as shown by multi-modal human pose estimation and human activity recognition and re-id tasks. 
%There is a natural correspondence between human activity data perceived, represented, or described through different data modalities, such as RGB cameras, LiDAR (Light Detection and Ranging), natural language, human skeletons, audio, or inertial measurement units (IMU). %Leveraging this correspondence, multi-modal fusion enhances various human activity understanding tasks, including human pose estimation, activity recognition, and person re-identification. 
%Despite recent works demonstrating the effectiveness of RGB videos as the primary visual modality for multi-modal representation learning, these models become impractical in scenarios where RGB cameras cannot be deployed, for instance, due to privacy concerns or lighting conditions. 
Despite LiDAR (Light Detection and Ranging) being an effective privacy-preserving alternative to RGB cameras to perceive human activities, it remains largely underexplored in the context of multi-modal contrastive pre-training for human activity understanding (e.g., human activity recognition (HAR), retrieval, or person re-identification (RE-ID)). To close this gap, our work explores learning the correspondence between LiDAR point clouds, human skeleton poses, IMU data, and text in a joint embedding space. More specifically, we present DeSPITE, a \underline{\textbf{D}e}ep \underline{\textbf{S}}keleton-\underline{\textbf{P}}ointcloud-\underline{\textbf{I}}MU-\underline{\textbf{T}}ext \underline{\textbf{E}}mbedding model, which effectively learns a joint embedding space across these four modalities. At the heart of our empirical exploration, we have combined the existing LIPD and Babel datasets, which enabled us to synchronize data of all four modalities, allowing us to explore the learning of a new joint embedding space. Our experiments demonstrate novel human activity understanding tasks for point cloud sequences enabled through DeSPITE, including Skeleton$\leftrightarrow$Pointcloud$\leftrightarrow$IMU matching, retrieval, and temporal moment retrieval. Furthermore, we show that DeSPITE is an effective pre-training strategy for point cloud HAR through experiments in MSR-Action3D and HMPEAR. 
\end{abstract}

\section{Introduction}
A key challenge in multi-modal human activity understanding tasks, such as human activity recognition (HAR), human pose estimation (HPE), retrieval, or person re-identification (RE-ID) ``in the wild'' is obtaining paired sensor data for each individual in a multi-person scene (e.g., IMU with human poses, point clouds, or RGB videos). Prior work has studied RGB-IMU matching for identity-aware tracking/RE-ID~\cite{huang2020wifi, cao2022vitag}, RGB-IMU matching for video retrieval~\cite{moon2023imu2clip}, and IMU-Skeleton Pose matching~\cite{vonMarcard2018} to correct IMU drift in multi-modal HPE. However, existing methods primarily focus on RGB-centric modalities, limiting applicability to privacy-sensitive scenarios like healthcare and surveillance, where RGB cameras may not be able to be deployed.

To address privacy concerns, silhouette masks~\cite{masullo2019goes, masullo2020person} or skeletons~\cite{bastico2022simultaneous} have been proposed to anonymize detected individuals from RGB video. %which then allows person re-identification through matching the IMU signals to the anonymized object representations. 
While effective, these anonymization techniques still come with the limitation that they require post-processing and short-term storage of the raw RGB data. In contrast, LiDAR is a privacy-preserving alternative, with proven capabilities for multi-modal HAR (e.g.,~\cite{xu2023human, lin2024hmpear}) and HPE (e.g.,~\cite{li2022lidarcap, ren2023lidar}). However, matching skeleton or IMU signals to LiDAR-based point cloud sequences is underexplored. 

Beyond matching, 
recent advances in multi-modal contrastive learning, such 
as ImageBind~\cite{girdhar2023imagebind}, IMU2CLIP~\cite{moon2023imu2clip}, BabelTower~\cite{dai2024advancing}, MotionCLIP~\cite{tevet2022motionclip}, or LAVIMO~\cite{yin2024tri} have demonstrated the power and benefits of cross-modal alignment for human activity understanding. These models learn a shared embedding space, enabling cross-modality matching, retrieval, and effective neural network pre-training for downstream tasks. Despite their success, they all use RGB data as a main visual modality to bind the learned representations. Extending this line of research, our paper asks the important research question \textit{What happens if we only depend on LiDAR in multi-modal contrastive learning as the main visual modality?}, which has not been studied before.

\begin{figure*}[t]
    \centering
    \includegraphics[width=0.8\linewidth]{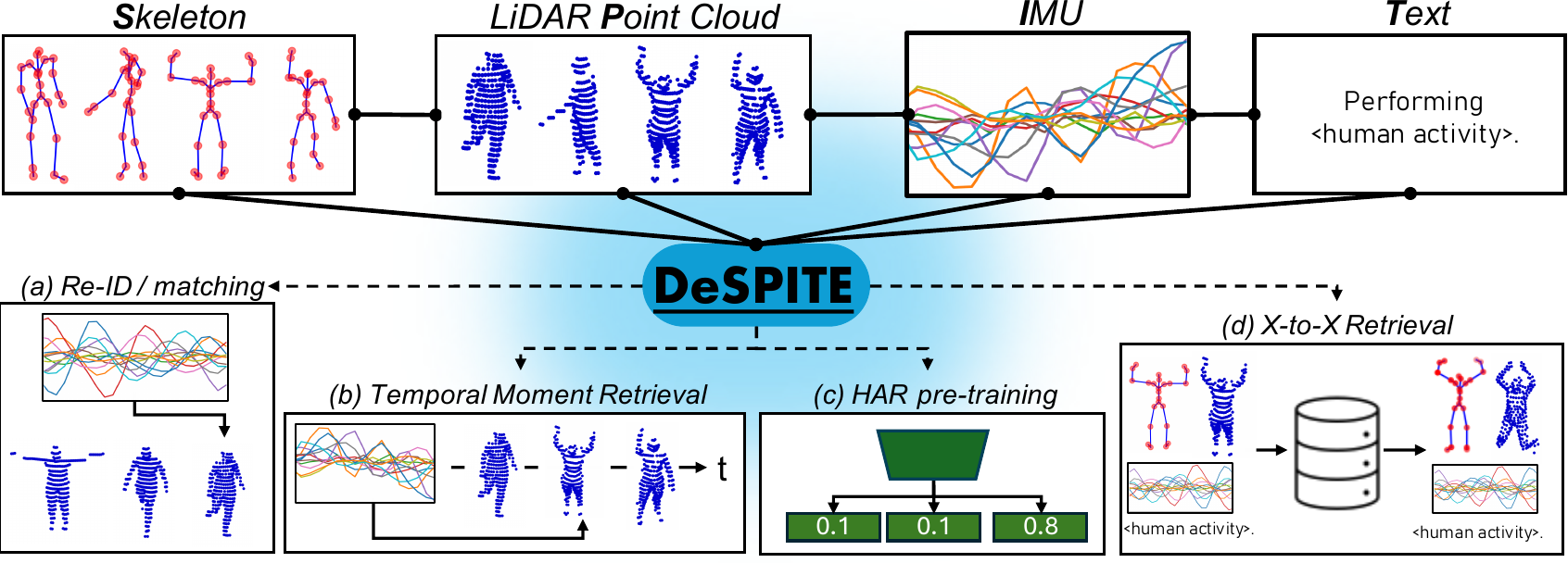}
    \caption{DeSPITE links different data modalities that represent human activities and, therefore, have a natural correspondence into a joint embedding space. As a result, DeSPITE enables tasks that depend on this correspondence that were not possible before.}
    \label{fig:overview}
\end{figure*}

We present DeSPITE, a \underline{\textbf{D}e}ep \underline{\textbf{S}}keleton-\underline{\textbf{P}}ointcloud-\underline{\textbf{I}}MU-\underline{\textbf{T}}ext \underline{\textbf{E}}mbedding model, which is illustrated in Figure~\ref{fig:overview}. Inspired by CLIP~\cite{radford2021learning}, ImageBind~\cite{girdhar2023imagebind}, and IMU2CLIP~\cite{moon2023imu2clip}, DeSPITE learns a shared embedding space with a contrastive loss based on InfoNCE~\cite{oord2018representation} between paired sequences of point~cloud$\leftrightarrow$skeleton$\leftrightarrow$IMU$\leftrightarrow$text data. Unlike prior works leveraging frozen text or RGB data embeddings as a binding modality (e.g., IMU2CLIP~\cite{moon2023imu2clip}, MotionCLIP~\cite{tevet2022motionclip}, LAVIMO~\cite{yin2024tri}), our primary goal is not to demonstrate modality alignment to text. Instead, we present novel applications for point cloud-based human activity sequences that were not possible before, enabled by unifying these modalities into a joint embedding space. Furthermore, while our primary focus is on enabling novel cross-modal retrieval and matching tasks, we find that DeSPITE also serves as an effective pre-training method for HAR, demonstrating improvements over uni-modal baselines. Finally, to understand the contribution of each modality to the joint embedding space, we train several DeSPITE variants (DeSPIE, DeSPE, etc.), evaluating their individual impact on cross-modal matching, retrieval, and HAR performance.
%Furthermore, we empirically show that pre-training a point cloud encoder in this way enhances HAR performance after fine-tuning, outperforming uni-modal pre-training. 
%To validate this, we train multiple DeSPITE variants (DeSIE, DeSPE, DePIE, DePITe, etc.), demonstrating comparable performance to DeSPITE and highlighting the potential of point cloud pre-training without implicit textual or RGB supervision.

As described in Figure~\ref{fig:overview}, after successful alignment in a shared embedding space, DeSPITE and
its variants (e.g., DeSPIE) allow novel and very useful applications between point cloud, IMU, and skeleton data: (a)~Person Re-ID by matching a snippet of, e.g., IMU data to the correct point cloud snippet, or a skeleton snippet to point cloud sequences, (b)~Temporal moment retrieval/search in a video with skeletons, IMU, or a point cloud as query, (c)~pre-training modality specific encoders for human activity recognition, and (d)~retrieval of each modality through each modality from a large motion database.

To train and evaluate DeSPITE, we construct LIPD-Babel, a dataset aligning point clouds, IMU, skeletons, and text by integrating the LIPD dataset~\cite{ren2023lidar} with the text annotations from Babel~\cite{BABEL:CVPR:2021}. LIPD-Babel enables two key evaluations: (1)~LIPD-Babel-v1 for cross-modal matching and retrieval, demonstrating DeSPITE and DeSPIE’s ability to align point clouds, skeletons, and IMU signals, and (2)~LIPD-Babel-v2 for contrastive pre-training, where DeSPITE and DeSPIE improve single-modality HAR, surpassing SOTA on HMPEAR~\cite{lin2024hmpear} and MSR-Action3D~\cite{li2010action}.

%To train and evaluate DeSPITE, we construct LIPD-Babel, a new dataset that aligns point clouds, IMU, skeletons, and text by integrating the LIPD~\cite{ren2023lidar} dataset, which provides synchronized point cloud, IMU, and skeleton data, with Babel~\cite{BABEL:CVPR:2021}, which extends AMASS~\cite{AMASS:ICCV:2019} with text annotations. LIPD-Babel enables rigorous evaluation of our approach in two key settings: (LIPD-Babel-v1) cross-modal matching and retrieval, demonstrating that DeSPITE and DeSPIE can successfully learn to match point clouds, skeletons, and IMU signals in the embedding space, and (LIPD-Babel-v2) contrastive pre-training for HAR, where we show that DeSPITE and DeSPIE improves single-modality HAR performance, outperforming state-of-the-art approaches on datasets such as HMPEAR~\cite{lin2024hmpear} and MSR-Action3D~\cite{li2010action}.

Our contributions are as follows. (i)~DeSPITE: A \underline{\textbf{D}e}ep \underline{\textbf{S}}keleton-\underline{\textbf{P}}ointcloud-\underline{\textbf{I}}MU-\underline{\textbf{T}}ext \underline{\textbf{E}}mbedding model that enables cross-modal matching and retrieval tasks between point clouds, IMU, and skeletons, unlocking applications that were previously impossible, and we will release the resulting pre-trained encoders, code, and data for future research. 
(ii)~We show that DeSPITE is an effective contrastive pre-training strategy for single-modality HAR, demonstrating new state-of-the-art performance. 
(iii)~LIPD-Babel, a new dataset for privacy-preserving multi-modal learning, with versions LIPD-Babel-v1 tailored for matching, retrieval tasks, and LIPD-Babel-v2 tailored for HAR.

\section{Related Work}

%Related work can be categorized into multi-modal contrastive representation learning, pre-training for point cloud human activity recognition, and LiDAR point cloud human activity datasets.

\subsection{Multi-Modal Contrastive Learning for Human Activity Understanding}
Recent works have explored unified embedding spaces across sensor modalities. ImageBind~\cite{girdhar2023imagebind} binds six modalities using image-text pairs, while IMU2CLIP~\cite{moon2023imu2clip} and MotionCLIP~\cite{tevet2022motionclip} align IMU and skeleton data with CLIP’s image-text space. LAVIMO~\cite{yin2024tri} improves skeleton-video-text retrieval, and BabelTower~\cite{dai2024advancing} incrementally aligns six sensing modalities, reducing reliance on RGB. Unlike these works, we focus exclusively on privacy-preserving modalities, introducing LiDAR into a joint embedding space with IMU and skeletons. This enables novel retrieval tasks (LiDAR$\leftrightarrow$IMU, LiDAR$\leftrightarrow$Skeleton) and serves as an effective pre-training strategy for point cloud-based HAR.

\subsection{Pre-Training for Point Cloud Human Activity Recognition}
While countless general purpose embedding models and foundation models have emerged in the last years for RGB images/videos, natural language, or audio (e.g, ImageBind~\cite{girdhar2023imagebind}, DinoV2~\cite{oquab2023dinov2}, SAM2~\cite{ravi2024sam2}, CLIP~\cite{radford2021learning}, BERT~\cite{devlin2019bert}), pre-trained general-purpose models for (LiDAR) point cloud HAR do not exist yet due to a lack of the same amount of data. To this day, pre-training mainly happens through self-supervision on a small number of datasets or the same dataset before fine-tuning for point cloud HAR. Self-supervision includes temporal order prediction from shuffling in~\cite{wang2021self, sheng2023point}, contrastive learning on masked sequences~\cite{shen2023pointcmp, han2024masked}, temporal structure prediction~\cite{shen2023masked}, or knowledge distillation~\cite{zhang2023complete}. All these methods have been shown to improve the performance after fine-tuning for HAR compared to training from scratch. Different from these works, we show that multi-modal contrastive learning between point clouds and other closely related modalities (i.e., skeleton pose and IMU data) leads to improved HAR performance after fine-tuning, showing new possibilities for future research in point cloud HAR.
% Mainly through self-supervision, this can drastically reduce the amount of data required to train on our own data, and for modalities like text or images there are countless pre-trained resources available due to the sheer amount of data. DinoV2, ResNet Imagenet, SAM2, Bert, Distillbert, CLIP, etc. 

\subsection{Cross-Modal Matching and Retrieval between Modalities}

Cross-modal matching assigns a data point from one modality to its correct counterpart in another. Key applications include audio-visual association~\cite{korbar2018cooperative, hamilton2024separating}, IMU-based matching to human pose, RGB, or silhouette masks~\cite{vonMarcard2018, moon2023imu2clip, masullo2019goes}, or text-to-motion retrieval~\cite{petrovich2023tmr, yin2024tri}. Person Re-ID via IMU signals has been explored in RGB videos~\cite{cao2022vitag, liang2024enhancing}, silhouette masks~\cite{masullo2019goes, masullo2020person}, and skeletons~\cite{bastico2022simultaneous, vonMarcard2018}. Retrieval tasks also exist between skeletons and text~\cite{petrovich2023tmr}, skeletons and RGB~\cite{yin2024tri}, and IMU and RGB~\cite{moon2023imu2clip}, with prior works exploring temporal moment retrieval and database retrieval.

However, LiDAR-based cross-modal retrieval remains largely unexplored. Our work extends these approaches by aligning LiDAR, IMU, and skeleton data, enabling novel retrieval tasks such as LiDAR$\leftrightarrow$Skeleton and LiDAR$\leftrightarrow$IMU. We further extend IMU interpretability via RGB video retrieval~\cite{moon2023imu2clip} to point cloud and skeleton retrieval, unlocking a new effective way to interpret IMU signals.

\subsection{Datasets for LiDAR Point Cloud Human Activity Recognition}

Early point cloud HAR datasets, like MSR-Action3D~\cite{li2010action} and NTU-RGB+D~\cite{shahroudy2016ntu, liu2020ntu} are derived from depth maps and have been foundational in advancing state-of-the-art methods in the field. Datasets with real LiDAR point clouds of human activities are rare. One of the only datasets for human activity recognition are HuCenLife~\cite{xu2023human}, and the recent HMPEAR~\cite{lin2024hmpear} and MM-Fi~\cite{yang2023mm} datasets. Motivated by multi-modal LiDAR and IMU-based HPE, several datasets have been proposed recently, such as LidarCap~\cite{li2022lidarcap} and LIPD~\cite{ren2023lidar}. In particular, LIPD is a large-scale dataset with human motions of LiDAR point clouds, human skeletons, and IMU data, but without activity annotations. It is a mix of synthetic and real data, where a big part comes from AMASS~\cite{AMASS:ICCV:2019}, a large-scale human motion capture dataset. On top of AMASS, Babel~\cite{BABEL:CVPR:2021} and HumanML3D~\cite{Guo_2022_CVPR} added natural language annotations. For our study, we combine LIPD with its corresponding subset in AMASS to a new dataset, LIPD-Babel, which enriches LIPD through partial human activity annotations. This leads to a large pre-training data resource between human skeletons, IMU, LiDAR point clouds, and text, which we can leverage to explore contrastive learning between these modalities.

\section{Method}

\subsection{Problem Statement}

The goal of this work is to learn a joint embedding space that aligns human motion observed through different privacy-preserving modalities. More specifically, we present DeSPITE, a \underline{\textbf{D}e}ep \underline{\textbf{S}}keleton-\underline{\textbf{P}}ointcloud-\underline{\textbf{I}}MU-\underline{\textbf{T}}ext \underline{\textbf{E}}mbedding model, which effectively learns a joint embedding space across these four respective modalities through a contrastive loss based on InfoNCE~\cite{oord2018representation}. 

We train several versions of DeSPITE, where we vary the number of modalities (DeSIE, DeSPE, DePIE, DePITE, ...). When all modalities are used, the text embeddings of CLIP serve as a binding modality, which has been shown to be effective in several recent related works for modalities that capture human motion, such as IMU data and human skeletons~\cite{girdhar2023imagebind, moon2023imu2clip, tevet2022motionclip, yin2024tri, bastico2022simultaneous, lu2024cross}. %Especially the works in~\cite{wu2023revisiting} and~\cite{moon2023imu2clip} have shown the effectiveness of binding several human motion data modalities to CLIP text embeddings.

%\textit{\textbf{*NOTE*}} 
We want to emphasize that the primary goal of DeSPITE is \textit{not} to show that we can bind skeleton, point cloud, or IMU data to CLIP text embeddings (this has been demonstrated before with, e.g., ImageBind~\cite{girdhar2023imagebind}, IMU2CLIP~\cite{moon2023imu2clip}, MotionCLIP~\cite{tevet2022motionclip}, or BabelTower~\cite{dai2024advancing}). Instead, our main goal is to %(i) empirically understand the impact of each modality in such a multi-modal binding framework for point cloud upstream and downstream tasks, and (ii) 
present several novel unexplored applications for human activity point cloud sequences that emerge when we unify these modalities into a joint embedding space.
%that have not been demonstrated before, that emerge when we unify these modalities into a joint embedding space. 
%Furthermore, we empirically show that pre-training a point cloud encoder in this way leads to stronger HAR performance after fine-tuning compared to uni-modal pre-training approaches.

\subsection{Learning \underline{\textbf{D}}eep \underline{\textbf{S}}keleton-\underline{\textbf{P}}ointcloud-\underline{\textbf{I}}MU-\underline{\textbf{T}}ext \underline{\textbf{E}}mbeddings (DeSPITE)}

Human motions represented through LiDAR point clouds, IMU time series, and human pose skeleton data have an inherent correspondence. We leverage this property to learn a joint embedding space where similar sequences of human motions are close and different sequences are far apart. 

Given a point cloud sequence $X_{pc} := \{ pc_{1}, \dots, pc_{T} \}$, with $pc_{i} \in \mathbb{R}^{N \times 3}$, an IMU sequence $X_{imu} := \{ imu_{1}, \dots, imu_{T} \}$, with $imu_{i} \in \mathbb{R}^{C}$, and a human pose sequence $X_{pose} := \{ pose_{1}, \dots, pose_{T} \}$, with $pose_{i} \in \mathbb{R}^{24 \times 3}$ representing 3D positions of 24 skeletal joints, we aim to train neural networks to encode $X_{pc}$, $X_{imu}$, and $X_{pose}$ into a shared embedding space. We denote these neural networks as encoders
\begin{align*}
f_{pc}: \mathbb{R}^{T \times N \times 3} \rightarrow \mathbb{R}^{e}, \\
f_{imu}: \mathbb{R}^{T \times C} \rightarrow \mathbb{R}^{e},        \\
f_{pose}: \mathbb{R}^{T \times 24 \times 3} \rightarrow \mathbb{R}^{e}
\end{align*}

which map the input sequences to embeddings \mbox{$z_{pc} = f_{pc}(X_{pc})$}, \mbox{$z_{imu} = f_{imu}(X_{imu})$}, and \mbox{$z_{pose} = f_{pose}(X_{pose})$}, where \mbox{$z_{pc}, z_{imu}, z_{pose} \in \mathbb{R}^e$}. Furthermore, we work with the setting where a natural language description $X_{text}$ is not provided for each respective $(X_{pc}, X_{pose}, X_{imu})$ triple. For this reason, the loss for text descriptions is only computed on the subset of the elements in each batch where we have a corresponding quadruple $(X_{pc}, X_{pose}, X_{imu}, X_{text}, tm)$, where $tm \in \mathbb{B}$ represents a boolean mask if there exists a text description $X_{text}$. In this way, we can effectively ignore the respective elements in the batch $B$ that do not have text descriptions when computing our alignment loss.

Following previous works like CLIP~\cite{radford2021learning}, ImageBind~\cite{girdhar2023imagebind}, MotionCLIP~\cite{tevet2022motionclip}, and IMU2CLIP~\cite{moon2023imu2clip}, we optimize our encoders using a contrastive objective based on InfoNCE~\cite{oord2018representation}. For a batch of $B$ paired samples, we obtain a boolean mask and embeddings ${ (z_{pc}^i, z_{imu}^i, z_{pose}^i, z_{text}^i, tm^{i})}_{i=1}^B$, where $z_{text}^i$ is obtained from a frozen CLIP text encoder. For batch elements $i$ without text pairings, we set $tm^{i}$ to 0 and use a dummy embedding, $tm^{i}$ is set to 1 otherwise. The similarity between embeddings is defined using the cosine similarity:

\begin{equation}
\mathrm{sim}(z_a, z_b) = \frac{z_a \cdot z_b}{\lVert z_a \rVert \lVert z_b \rVert},
\end{equation}

where $z_a, z_b \in \{ z_{pc}, z_{imu}, z_{pose}, z_{text} \}$. The contrastive loss for each pair $(i, j)$ in the batch is defined as follows:

\begin{equation}
\mathcal{L}_{a \to b}^i = - \log \frac{\exp(\mathrm{sim}(z_a^i, z_b^i) / \tau)} 
{\sum_{j=1}^B \exp(\mathrm{sim}(z_a^i, z_b^j) / \tau)}
\end{equation}

where $a, b \in \{ pc, imu, pose, text \}$ and $\tau > 0$ is a (learnable) temperature hyperparameter. Symmetrically, we compute the loss in both directions by swapping the roles of the modalities, i.e., $\mathcal{L}_{a \to b}^i$ and $\mathcal{L}_{b \to a}^i$, which leads to:

\begin{equation}
\mathcal{L}_{a, b}^i = \dfrac{1}{2}(\mathcal{L}_{a \to b}^i + \mathcal{L}_{b \to a}^i)
\end{equation}

%For all elements in the batch, we compute the bidirectional losses for each pair of modalities (point cloud-IMU, point cloud-pose, IMU-pose, point cloud-text, pose-text, IMU-text). 

As our main goal is to align $z_{pc}, z_{imu}, z_{pose}$, we employ two different losses. First, we bind the subset of paired $z_{pc}, z_{imu}, z_{pose}$ with the respective text embeddings $x_{text}$:

\begin{equation} 
\mathcal{L}_{text}^{i} = \sum_{i=1}^{B} tm^{i} \sum_{a \in M} \mathcal{L}_{a, text}^i%\dfrac{1}{2} (L_{a, text} + L_{text, a}) 
\end{equation}
where $tm^{i}$ serves as a mask to ignore the elements in the batch without text pairings for this loss. Second, each individual sensing modality pair \mbox{$M^{*} := \{ (pc, imu), (pc, pose), (imu, pose) \}$} is optimized to be close to each other: 

\begin{equation} \label{eq:alignment_loss}
\mathcal{L}_{M}^{i} = \sum_{i=1}^{B} \sum_{ (a,b) \in M^{*}} \mathcal{L}_{a, b}^i 
%\dfrac{1}{2}(L_{a,b} + L_{b,a})
\end{equation}

In both $\mathcal{L}_{text}$ and $\mathcal{L}_{M}$, we do not weight each modality individually. Finally, we combine both losses to enforce aligning embeddings from the corresponding point cloud, IMU, and pose sequences while constraining them to take small steps toward the text embedding space of CLIP. With \mbox{$M := \{pc, imu, skeleton\}$} being the set of modalities to align and $M^{*}$ their respective desired pairings, we optimize the following final loss function for each batch:

\begin{equation}
\mathcal{L}_{total}^{i} = \alpha \mathcal{L}_{text} + \beta \mathcal{L}_{M}
\end{equation}
where $\alpha=0.5, \beta=0.5$ equally weight both loss terms.

In our experiments, we train models of all possible modality combinations, which requires an according change to the modality set $M$ and the respective pairings $M^{*}$ (e.g., training only DeSPE, then $M := \{skeleton, pointcloud\}$). Finally, when training a model like DeSPE without text pairings, the overall loss simplifies to Equation~\ref{eq:alignment_loss}, so that
\mbox{$\mathcal{L}_{total}^{i} = \mathcal{L}_{M}$}. 

%This loss encourages aligned embeddings from the corresponding point cloud, IMU, human pose sequences, and text while pushing apart non-matching pairs in the batch. 

%This has two advantages, (i) The matching from one modality to the other becomes more robust given that several sequences are observed, (ii) the whole method becomes more robust to occlusions given that matching is recalculated all the time. 
%We present an effective matching algorithm shown in Algorithm~\ref{alg:1}, where the mean similarity score over several embeddings from a short temporal neighborhood determines the mapping. More specifically, given source modality $a$ and target matching modality $b$, we compute the average over the similarity score between $N$ consecutive encoding time points. 

\section{Experiments}

We evaluate the effectiveness of DeSPITE and its variants on the following tasks: Modality matching, temporal moment retrieval using a different modality as a query, pre-training for point cloud human activity recognition, and several qualitative evaluations.
%To this end, we use 

%by constructing a large-scale Skeleton-Pointcloud-IMU-Text dataset, which is a combination of the LiPD and Babel dataset. With these datasets, we evaluate the effectiveness of matching, temporal localization, retrieval, and human activity recognition. In addition, we evaluate the performance for pre-trainin for HAR tasks on MSRAction3D and HMPEAR.

\subsection{Datasets}
We train our method on a merged version of LIPD~\cite{ren2023lidar} and Babel~\cite{BABEL:CVPR:2021} (denoted as Babel+LIPD), where we map the text annotations from Babel to the AMASS~\cite{AMASS:ICCV:2019} subsets present in LIPD. In this way, we are able to construct a large-scale dataset of real and synthetic LiDAR point cloud, IMU, and skeleton data with text annotations. To be more specific, we construct two versions of LIPD-Babel. First LIPD-Babel-v1, where we use the official train-test split of LIPD\footnote{We exclude LidarCap~\cite{li2022lidarcap} since the data is not available in the official LIPD data repository. More details about LIPD-Babel in Appendix~\ref{app:dataset}.}, including DIP~\cite{huang2018deep} and TotalCapture (TC)~\cite{trumble2017total}. Second LIPD-Babel-v2, where we use the train-val split of Babel for the AMASS subsets, and add all the remaining data of LIPD to the training set. As LIPD is provided in 10 FPS, we downsample the Babel annotations to 10 FPS. After preprocessing the whole dataset with sliding windows of length 24, we obtain 502,958 / 85,551 training/testing windows for LIPD-Babel-v1, from which 85,551 training windows have text annotations, and 403,430 / 58,802 train/test windows for LIPD-Babel-v2, with 135,699 text training windows and 58,802 test annotations.  % accordingly. %Notably, we do not merge text annotations for the LIPD-Babel-v1 test set, as this version of the dataset is not intended for HAR. 
%On the first version, ``LIPD-Babel-v1'', we evaluate all our matching and temporal retrieval tasks on three challenging unseen tests (eLIPD, eTC, eDIP). On the second version, ``LIPD-Babel-v2'', we can evaluate text-to-motion retrieval against different state-of-the-art methods based on the subset that we have. Furtermore, we can evaluate zero-shot HAR on pointclouds and IMU. Finally, point cloud to skeleton and imu to skeleton retrieval is evaluated qualitatively in LIPD-Babel-v1 and quantitatively in LIPD-Babel-v1, where the text labels of the respective test set serve as the retrieval ground truth.

Regarding downstream task performance for HAR, we evaluate our approach on HMPEAR~\cite{lin2024hmpear}, MSR-Action3D~\cite{li2010action}, and our Babel-LIPD-v2 train/test split that only includes Babel sequences. Both HMPEAR and MSR-Action3D include domain shifts, where HMPEAR uses a different kind of LiDAR sensor, and MSR-Action3D has very dense point clouds derived from depth maps. %In this way, our evaluation effectively allows us to test the pre-training capabilities of DeSPITE for several target domains.
%\paragraph{HMPEAR}

%\paragraph{MSR-Action3D}

\subsection{Experimental Design and Metrics}
%Our experiments are designed to show that the correspondence between the modalities included in DeSPITE can be learned effectively. 
We use the following tasks to evaluate the performance of DeSPITE (and its variants) and enable future research to compare against our baselines. Throughout all models in our experiments, all hyperparameters are kept the same. % (e.g., embedding dimension, number of layers, number of epochs, learning rate, sequence length, ...). % so that any difference in performance can only be reduced to the respective training modalities. 

%\subsection{Experimental Design}
%Our experiments are designed to show that the correspondence between the modalities included in SPITE can be learned effectively. To this end, the correspondence between each modality is evaluated through a large-scale ablation study, where we carefully ablate all combinations of the respective modalities. Throughout all our experiments, all training and model hyperparameters are kept the same (e.g., embedding dimension, number of layers, number of epochs, learning rate, sequence length, ...) so that any difference in performance can only be reduced to the respective training modalities. 

%Given that our model is the first of its kind to explore the correspondence between these modalities, it is not possible to directly compare against other state-of-the-art approaches. We know the recent work proposed in~\cite{dai2024advancing}, where IMU, skeleton, and point cloud data are aligned among other sensing modalities. However, their code has not been made publicly available, so that a direct comparison against their approach is not possible in our large-scale ablation study. Furthermore, our model differs in small detail. For all the data we use, there is a matching pair between all modalities (except for text). 

\subsubsection*{Task 1. Matching between Modalities}
In multi-person scenes, matching IMU data to detected individuals in point cloud sequences is a challenging upstream task, which has not been explored before. This task can be generalized to an any-to-any modality matching problem, which we even further evaluate with this task. We evaluate all modality combinations IMU$\leftrightarrow$PC, IMU$\leftrightarrow $Skeleton, and PC$\leftrightarrow$Skeleton. %DeSPITE’s joint embedding space enables modality matching for downstream applications ``in the wild'' such as multi-modal HAR, HPE, and identity-aware tracking. We evaluate matching between all modality combinations on three held-out test sets.
For each test set (LIPD-Test, TC, DIP), we generate 1000 artificial multi-person scenes (following designs in prior works~\cite{masullo2019goes, mukashev2022person}). This is achieved by randomly sampling $n$ sequences from the test set first and then sampling a respective subsequence, leading to $n$ artificial subjects carrying out an activity simultaneously. The number of subjects per scene varies $n \in (2,4,8,12,16,20,24,28,32)$, simulating different real-world scenarios. Given $n$ subjects, we report matching accuracy through argmax on the cosine similarities per row between all candidates.  %This leads to 216,000 total matching evaluations per testing dataset. %Further, we vary the number of matching windows for Algorithm~\ref{alg:1}, i.e., $(1,2,3,4)$ consecutive windows, resulting in 36,000 total matching evaluations per dataset.

%Pointcloud$\leftrightarrow$IMU: Multi-person scenes may include several individuals who willingly share their IMU data~\cite{masullo2020person, cao2022vitag} recorded from several wearable devices attached to their body. Mapping this data to detected individuals (through, e.g., an object detection or instance segmentation+tracking method) in the scene is a non-trivial task. As an upstream task, the joint embedding space of DeSPITE effectively allows obtaining a matching between each modality associated with an individual in the scene for further downstream tasks (e.g., multi-modal HAR/HPE or identity-aware tracking). The performance is measured on each held-out test set. To carefully evaluate this task, we sample 1000 randomly generated artificial scenes (as done in previous works by~\cite{masullo2019goes, mukashev2022person}) to evaluate matching performance. Furthermore, we vary the number of subjects in each artificial scene, i.e., $(2,4,8,12,16,20,24,28,32)$, to simulate different real-world situations closely. Further, we vary the number of matching windows for Algorithm~\ref{alg:1}, i.e., $(1,2,3,4)$ consecutive windows. In total, our evaluation protocol yields 1000*9*4=36000 total matching performance results for each dataset.

\subsubsection*{Task 2. Temporal Moment Retrieval between Modalities}
Given a short snippet in one modality, the goal is to retrieve the corresponding temporal moment in the sequence observed with another modality. This task has been explored for, e.g., IMU-RGB~\cite{moon2023imu2clip} and skeleton-text~\cite{petrovich2023tmr}, but not yet for LiDAR point clouds, IMU, and skeletons. %combined through DeSPITE (or its variants, such as DeSPIE). 
We evaluate this on the three held-out test sets of LIPD (LIPD-Test, TC, DIP) using Recall@k $(k=1,10,20,50)$ shots across all modality combinations. Performance is measured by computing the cosine similarity scores for \textit{all} possible query-target pairs in \textit{all} individual test set sequences and returning the top-k similar frame indices. For each query, we compute the difference between all top-k returned time points against the ground truth. A retrieval is considered to be correct if it is within 10 frames $(\sim1.5 sec)$ of the ground truth. As the final score, the mean over all recall@k scores of all sequences for a dataset is reported.
%Given a short snippet of one embedding, we can retrieve or in other words, search for the specific point in time in another modality.In our experiment, a temporal retrieval is considered to be correct if it is at most 10 frames different from the ground truth (which is approximately half of the window’s size, which is roughly 1.5 seconds). We measure the performance on the held-out test set using Recall with $k \in (1,10,20,50)$ shots for all possible modality combinations and all models. More specifically, for each individual sequence in the test set, we compute the similarity between each possible pair of query and target windows, giving a precise estimation of the performance across an enormous number of sequences.

%Pointcloud$\leftrightarrow$IMU, IMU$\leftrightarrow$Skeleton, and Pointcloud$\leftrightarrow$Skeleton Retrieval: Matching between skeletons, point clouds, IMUs allows to analyze motion signals based on different representations. While IMU signals are hard to interpret in nature, the annotation process may become easier if a corresponding skeleton or point cloud representation can be visualized. The same holds for point clouds, where a skeleton representation may be easier to interpret. Further, given that large-scale skeleton datasets exist that may also be paired with text or other modalities, transitive retrieval of labels for the corresponding pointcloud or IMU signals will be possible in future work.

\subsubsection*{Task 3. Pre-Training for Human Activity Recognition}

We evaluate cross-modal pre-training for point clouds, IMUs, and skeletons via linear/non-linear probing and fine-tuning. HAR pre-training/testing is done on LIPD-Babel-v2, with additional point cloud testing on HMEPAR and MSR-Action3D. Results follow standard metrics: clip segment accuracy for MSR-Action3D, segment accuracy for HMEPAR and LIPD-Babel-v2 (excluding transition labels). We do not evaluate with additional skeleton/IMU datasets, since transfer learning is strongly limited by serious dataset-specific variations in joints and different IMU channel counts for these modalities. %Fas prior works (e.g., MotionCLIP+\cite{tevet2022motionclip}, IMU2CLIP+\cite{moon2023imu2clip}) already validated cross-modal pre-training.

%We assess cross-modal pre-training for point clouds, IMUs, and skeletons, using linear/non-linear probing and finetuning. HAR pre-training and testing are performed in LIPD-Babel-v2, while testing is additionally evaluated on HMEPAR and MSR-Action3D for point clouds. MSR-Action3D results are reported using clip segment accuracy, as per standard practice. For HMEPAR and LIPD-Babel-v2, we report segment accuracy, excluding transition activities in the latter. We do not evaluate other datasets for skeletons and IMU, previous works (e.g., MotionCLIP+\cite{tevet2022motionclip}, IMU2CLIP+\cite{moon2023imu2clip}) have already shown the effectiveness of cross-modal pre-training. Further, the representations between datasets vary significantly, which prohibits re-using the features learned for transfer learning between IMU datasets with varying numbers of channels or joint positions and skeleton datasets with different numbers of joints and their positions in the skeleton.

\subsubsection*{Task 4. Retrieval between Modalities from Database}
We qualitatively evaluate retrieval from a ``large database'' between Point Cloud$\leftrightarrow$IMU, IMU$\leftrightarrow$Skeleton, and Point Cloud$\leftrightarrow$Skeleton. This enables motion analysis across representations, aiding interpretability (e.g., skeletons or point clouds simplify IMU visualization). %Additionally, transitive retrieval can leverage large-scale datasets paired with other modalities for future research.
%Cross-modal pre-training can serve as an effective pre-training strategy for point clouds, IMU, and skeleton, which requires linear or linear probing, or transfer learning on a different dataset. We use the LIPD-Babel-v2 test split to evaluate pre-training for HAR, while also evaluating transfer learning to HMEPAR and MSR-Action3D for point clouds. For MSR-Action3D we report clip segment accuracy, since this is the commonly used setup among recent works. For HMPEAR, we only report segment accuracy given that the evaluation procedure for clips has not been made publicly available. For LIPD-Babel-v2 we also report segment accuracy only, but we ignore the ``transition'' activity class.

%\subsubsection*{Task 4. Retrieval with Text}
%To the best of our knowledge, retrieval of point cloud sequences with text is an underexplored field. Using DeSPITE, we offer a point cloud encoder that is paired with CLIP, so that zero shot HAR can become possible in the future for point cloud sequences as well. At the same time, all our modalities can be queried through text.

\subsection{Implementation Details}

For point clouds, we use the PST-Transformer~\cite{fan2022point} with a SimCLR-based projection head~\cite{chen2020simple}. IMU is encoded with a 2-layer LSTM~\cite{hochreiter1997long}, skeletons with the ACTOR encoder~\cite{petrovich2021action}, and text with a frozen CLIP text encoder~\cite{radford2021learning}. All models are pre-trained for 145 epochs with 512-d embeddings, Adam optimizer~\cite{kingma2014adam}, lr=1e-4, batch size 1024. We subsample 256-points using farthest point downsampling (FPD) on each frame and use 24-frame windows as input to all models. Augmentations (random translation, scaling, Gaussian noise) are employed during training to prevent overfitting. For a fair comparison, we only use the weights from epoch 145 across all models. HAR fine-tuning roughly follows~\cite{fan2022point}, with batch size 24, 35 epochs (SGD~\cite{ruder2016overview}, warmup to lr=0.01, 0.1 decay at epochs 20, 30). In HMPEAR, we subsample 1024 points using FPD and use 24-frame windows. In MSR-Action3D, we follow the standard 2048-point, 24-frame window setting.

\subsection{Results: Multi-Person LiDAR-IMU Matching}
\begin{figure}[t]
    \centering
    \includegraphics[width=\linewidth]{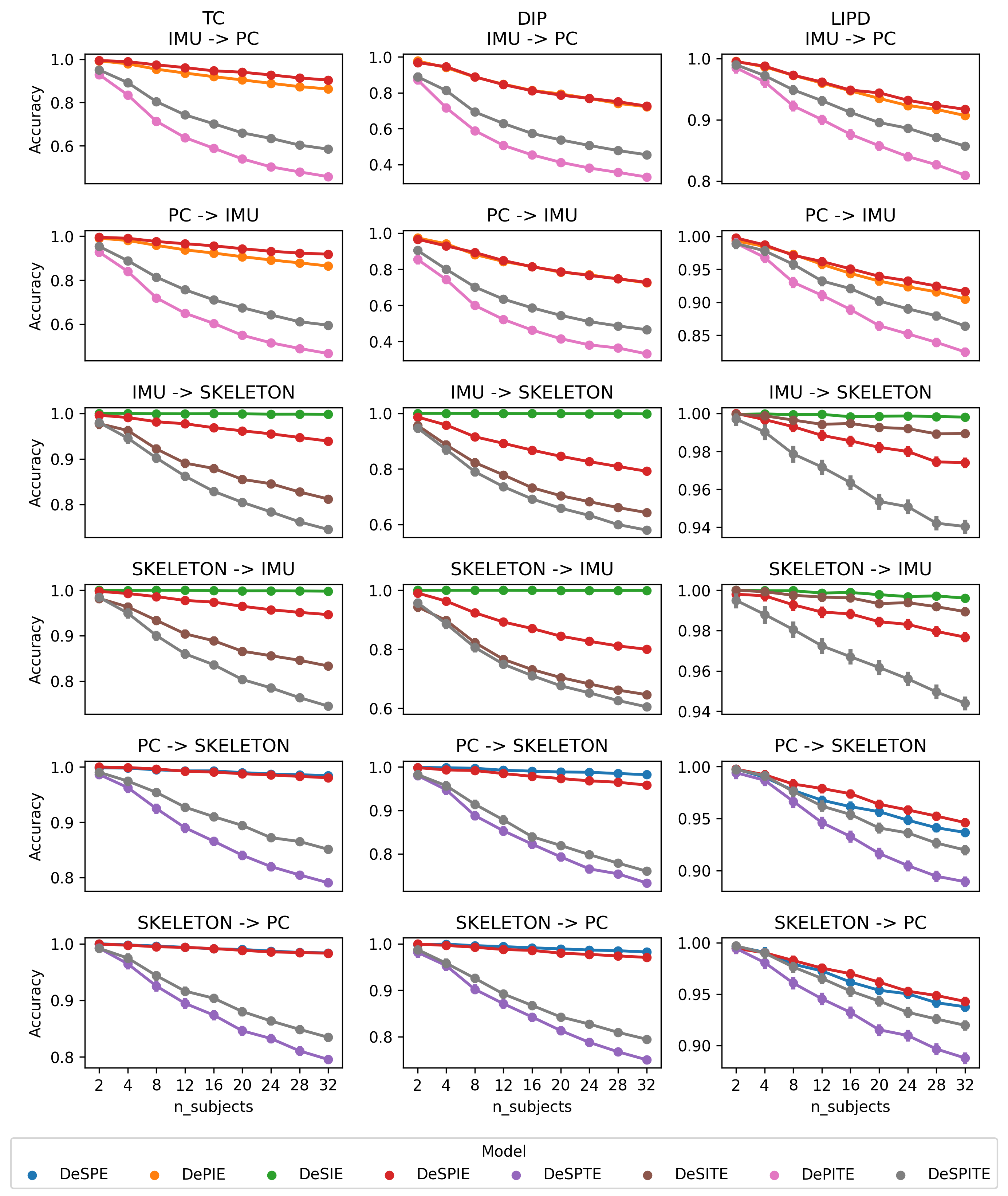}
    \caption{Matching performance between all modality pairs IMU$\leftrightarrow$Pointcloud, IMU$\leftrightarrow$Skeleton, Pointcloud$\leftrightarrow$Skeleton, reporting mean accuracy for $n\in(2,4,8,12,16,20,24,28,32)$ subjects across 1000 artificial scenes.}
    \label{fig:matching_performance}
\end{figure}

Figure~\ref{fig:matching_performance} shows our results for matching between IMU$\leftrightarrow$PC, IMU$\leftrightarrow $Skeleton, and PC$\leftrightarrow$Skeleton across all trained model variants (all specific numbers in Appendix~\ref{app:scores}). The subjects are varied on the x-axis, and matching accuracy is reported on the y-axis. Each row corresponds to the respective test set (TC, DIP, LIPD). First, our experiments reveal that models trained with text (i.e., DeSPITE, DePITE, DeSITE, DeSPTE) in almost all scenarios perform worse than models trained solely on the modalities alone (i.e., DeSPE, DePIE, DeSIE, DESPIE), showing that this task does not benefit from text embeddings. Second, we find that matching between IMU, point clouds, and skeletons can be effectively learned, showing up to perfect matching scores for a smaller number of subjects. In comparison, a larger number of subjects, as expected, becomes more challenging.

%Using our matching algorithm, we evaluate matching performance among random IMU sequences to LiDAR sequences. To this end, we construct artificial scenes, which has been done previously by~\cite{masullo2019goes, mukashev2022person} to evaluate the same task with silhouettes. We compare our approach against a randomly initialized model to emphasize the effectiveness of contrastive association due to lack of related works as we are the first to propose this task. As we are the first to propose this task for IMU and LiDAR, there are no direct competitors that allow us to perform a quantitative evaluation. We set the current SOTA with our approach and pave the way for future research in this domain. For the sake of comparison, we compare our results against the results of vision-based approaches, which in itself is not a fair comparison.

\subsection{Results: Temporal Moment Retrieval}
\begin{figure}[ht]
    \centering
    \includegraphics[width=\linewidth]{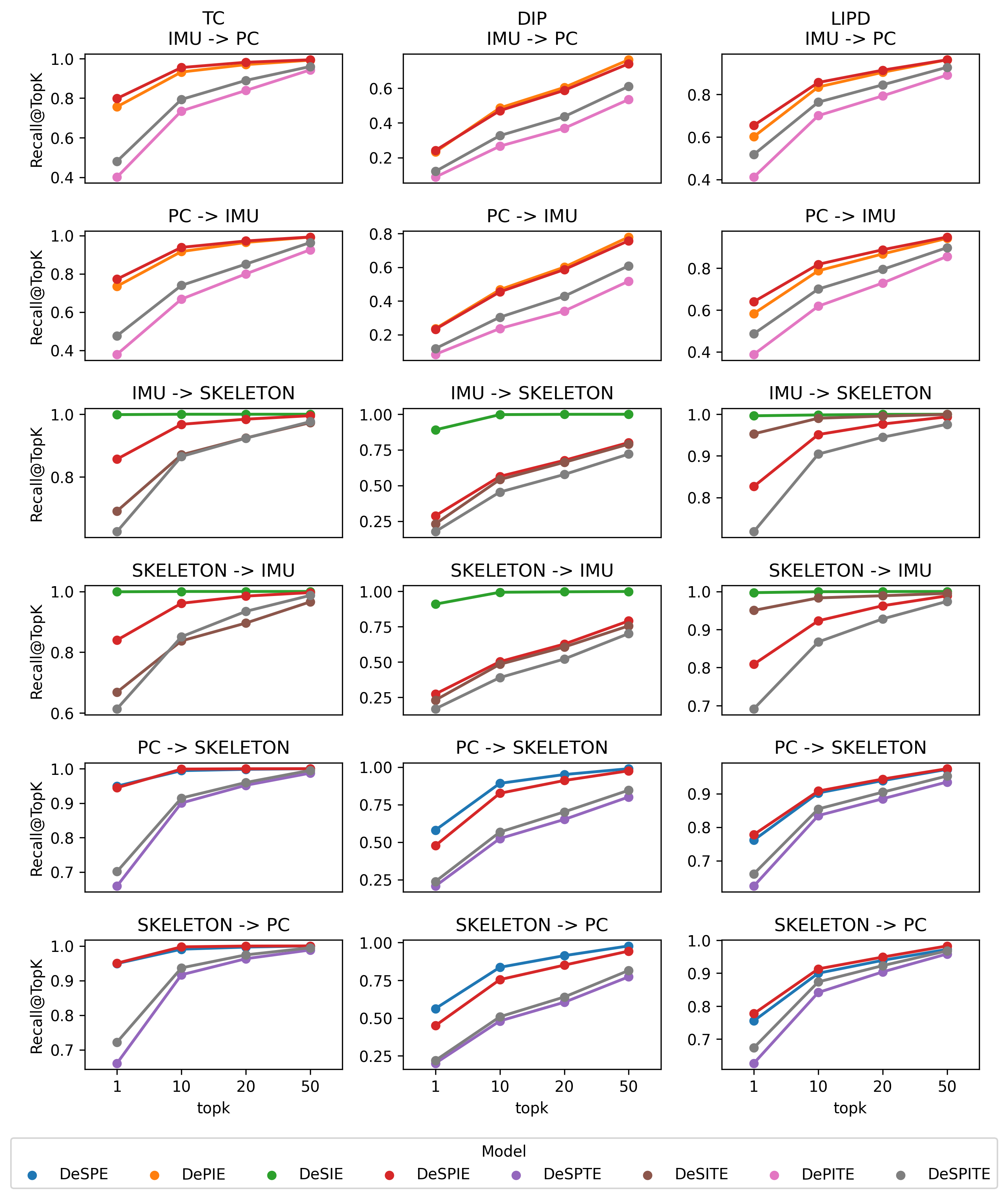}
    \caption{Temporal moment retrieval performance across all modalities. Recall is reported for top-1, 10, 20, and 50 retrievals, considering a match correct if within 10 frames of the ground truth ($\approx$half the window size). }
    \label{fig:moment_retrieval}
\end{figure}

Figure~\ref{fig:moment_retrieval} shows our results for temporal moment retrieval between IMU$\leftrightarrow$Pointcloud, IMU$\leftrightarrow$Skeleton, and Pointcloud$\leftrightarrow$Skeleton across all trained model variants (all specific numbers in Appendix~\ref{app:scores}). The $k$ retrieval shots are varied on the x-axis, and the respective recall@k is reported on the y-axis. Each row shows the results for each respective test set (TC, DIP, LIPD). First, we observe the same result for temporal moment retrieval as for matching in Figure~\ref{fig:matching_performance}: All models trained with text perform worse than models trained solely on the modalities alone. Second, our evaluation demonstrates that temporal moment retrieval can be solved the best between IMU$\leftrightarrow$Skeleton, where DeSIE demonstrates that training between both modalities alone is very effective. The runner-up is Pointcloud$\leftrightarrow$Skeleton where DeSPIE and DeSPE achieve almost identical performance. Finally, our experiments reveal that the most challenging problem is IMU$\leftrightarrow$Point cloud matching, allowing future work to propose more effective solutions.

\subsection{Results: 3D Point Cloud Human Activity Recognition}
The pre-trained embeddings of all versions of DeSPITE can be fine-tuned for HAR. We compare our approach against the recent state-of-the-art on MSR-Action3D, HMPEAR, and perform ablations on the LIPD-Babel-v2 split. 

%\textbf{MSR-Action3D:} %The results for MSR-Action3D are presented in Table~\ref{tab:msr_action3d_24frame_results}. We compare our method against all recent works, including several uni-modal point cloud pre-training approaches that pre-train already on MSR-Action3D. In comparison, fine-tuning our pre-trained embeddings from DeSPITE, DeSPIE, or DePITE outperforms all other pre-training methods despite encountering a severe domain shift from 256 points in the pre-training stage to 2048 points when fine-tuning. Our pre-training method together with PST-Transformer even outperforms the recently released PvNext (94.77 $<$ 95.47) and MAMBA4D (93.38 $<$95.47) approaches, which include notable architecture improvements. We even achieved almost the same performance as KAN-HyperpointNet, which is currently state-of-the-art (95.59 vs. our 95.47).

\textbf{MSR-Action3D:} Table~\ref{tab:msr_action3d_24frame_results} shows that fine-tuning DeSPITE, DeSPIE, or DePITE embeddings surpasses \textit{all} current state-of-the-art point cloud HAR pre-training methods, despite encountering a domain shift from 256 to 2048 points. Our approach, combined with PST-Transformer, even outperforms PvNext~\cite{wangpvnext} (94.77$<$95.47) and MAMBA4D~\cite{liu2024mamba4d} (93.38$<$95.47) and nearly matches KAN-HyperpointNet~\cite{chen2024kan} (95.59$>$95.47).

\textbf{HMPEAR:} 
As shown in Table~\ref{tab:hmpear_results}, we achieve new SOTA on HMPEAR, outperforming all prior point cloud, RGB, and multi-modal approaches. While our setup uses twice the frames of previous methods, pretraining PST-Transformer in the same setup with DeSPITE, DeSPIE, or DePITE improves its performance by nearly 4\%, demonstrating the effectiveness for HAR pre-training.

%, demonstrating the effectiveness of contrastive cross-modal pre-training.
%The results for HMPEAR are presented in Table~\ref{tab:hmpear_results}. We achieve new state-of-the-art on HMPEAR, outperforming all previously reported point cloud, RGB, and multi-modal approaches. Given that the training procedure and specific setup for the point cloud approaches has not been reported, we have to note that compared to the previous methods, we use double the amount of frames. However, we retrain the PST-Transformer on our version of the HMPEAR dataset, and we can see that pre-training it with DeSPITE, DeSPIE, or DePITE effectively improves the performance of PST-Transformer by almost 4\%, showing the strong benefit of contrastive cross-modal pre-training. 

\textbf{LIPD-Babel-v2:} 
Table~\ref{tab:babelcls_results} shows that all our models outperform baselines (PST-Transformer, LSTM, ACTOR) when trained from scratch on LIPD-Babel-v2. We explore various freezing strategies, as well as linear/non-linear probing and projection heads, with detailed ablations in the supplementary material (Tables~\ref{tab:babel_imu},\ref{tab:babel_pc},\ref{tab:babel_skeleton}). In Table~\ref{tab:babelcls_results}, only the best results of DePITE, DeSPIE, and DeSPITE are reported, which consistently achieve strong performance across all three datasets.

Notably, across MSR-Action3D and HMPEAR, DeSPITE, DeSPIE, and DePITE consistently achieve the best performance, underlining the advantage of pre-training with more modalities. Furthermore, different from the results for matching and temporal moment retrieval, we find that training with text benefits the fine-tuning performance for HAR.

%The results for LIPD-Babel-v2 are presented in Table~\ref{tab:babelcls_results}. All our models improve the results on this dataset compared to training the baselines (PST-Transformer, LSTM, and ACTOR) for each respective modality from scratch. We test all possible combinations with freezing the weights as well. Detailed ablations between all modalities, linear and non-linear probing, as well as fine-tuning with a linear and non-linear projection head are provided in the supplementary material (see Tables~\ref{tab:babel_imu},~\ref{tab:babel_pc},~\ref{tab:babel_skeleton}).

%Notable, for both MSR-Action3D and HMPEAR, the same three models (DeSPITE, DeSPIE, or DePITE) achieve the best performance, which shows that training with three modalities consistently shows the best performance.

% Requires: \usepackage{multirow}

\begin{table}[t]
    \centering
    \resizebox{0.8\columnwidth}{!}{%
    \begin{tabular}{lcc}
        \toprule
        \textbf{Methods} & \textbf{Video Acc@1 ($\uparrow$}) \\
        \midrule
        \multicolumn{2}{c}{\textit{Supervised Learning Only}} \\
        \midrule
        MeteorNet~\cite{liu2019meteornet} & 88.50 \\
        PSTNet~\cite{fan2021pstnet} & 91.20 \\
        P4Transformer~\cite{fan2021point} & 90.94 \\
        Kinet~\cite{zhong2022no} & 93.27 \\
        PPTr~\cite{wen2022point} & 92.33 \\
        PSTNet++~\cite{fan2021deep} & 92.68 \\
        Leaf~\cite{liu2023leaf} & 93.84 \\
        PST-Transformer~\cite{fan2022point} & 93.73 \\
        $\text{PST-Transformer}^{\dagger}$~\cite{fan2022point} & 92.33 \\ %94.42 \\
        MAMBA4D~\cite{liu2024mamba4d} & 93.38 \\
        PvNext~\cite{wangpvnext} & 94.77 \\
        KAN-HyperpointNet~\cite{chen2024kan} & 95.59 \\
        \midrule
        \multicolumn{2}{c}{\textit{Uni-Modal Pre-Training + Transfer Learning}} \\
        \midrule
        PSTNet + PointCPSC~\cite{sheng2023point} & 92.68 $(+1.48)$\\
        PSTNet + PointCMP~\cite{shen2023pointcmp} & 93.27 $(+2.07)$ \\
        PST-Transformer + MaST-Pre~\cite{shen2023masked} & 94.08 $(+0.35)$\\
        PPTr + C2P~\cite{zhang2023complete} & 94.76 $(+2.43)$ \\
        P4Transformer + M2PSC~\cite{han2024masked} & 93.03 $(+2.09)$ \\
        PST-Transformer + M2PSC~\cite{han2024masked} & 94.84 $(+1.11)$ \\
        \midrule
        \multicolumn{2}{c}{\textit{Multi-Modal Pre-Training + Transfer Learning}} \\
        \midrule
        PST-Transformer + \textbf{DePITE (Ours)} & \textbf{95.12} $(+1.39 \ \ | \  +2.79^\dagger)$ \\
        PST-Transformer + \textbf{DeSPIE (Ours)} & \textbf{95.47} $(+1.74 \ \ | \ +3.14^\dagger)$  \\
        PST-Transformer + \textbf{DeSPITE (Ours)} & \textbf{95.47} $(+1.74 \ \ | \ +3.14^\dagger)$ \\
        %PST-Transformer + CLIA (Ours) & 93.37 \\
        %PST-Transformer + \textbf{CLIA-XA (Ours)} & \textbf{95.12} \\
        \bottomrule
    \end{tabular}%
    }%
    \caption{24-frame classification results on the MSR-Action3D dataset, clip-level accuracy (Acc) is reported.}
    \label{tab:msr_action3d_24frame_results}
\end{table}

\begin{table}[t]
    \centering
    \resizebox{0.8\columnwidth}{!}{%
    \begin{tabular}{lc|c}
        \toprule
        Method & Modality & Acc(Seg)↑ \\
        \midrule
         \multicolumn{3}{c}{\textit{Uni-Modal Supervised Learning Only}} \\
        \midrule
        PSTNet~\cite{fan2021pstnet} & PC & 64.3 \\
        P4-Transformer~\cite{fan2021point} & PC  & 63.9 \\
        $\text{PST-Transformer}^{\dagger}$~\cite{fan2022point} & PC & 65.94 \\
        \midrule
        I3D~\cite{carreira2017quo} & RGB & 55.5 \\
        SlowFast~\cite{feichtenhofer2019slowfast} & RGB  & 62.2 \\
        TimeSformer~\cite{bertasius2021space} & RGB  & 56.3 \\
        Uniformer~\cite{li2022uniformer} & RGB  & 61.6 \\
        \midrule
        \multicolumn{3}{c}{\textit{Multi-Modal Supervised Learning Only}} \\
        \midrule
        AR-Proj~\cite{lin2024hmpear} & RGB+PC  & 60.6 \\
        PEAR-Proj (BestPE)~\cite{lin2024hmpear} & RGB+PC  & 64.1 \\
        PEAR-Proj (BestAR)~\cite{lin2024hmpear} & RGB+PC & 66.0 \\
        \midrule
        \multicolumn{3}{c}{\textit{Multi-Modal Pre-Training + Transfer Learning}} \\
        \midrule
        PST-Transformer + \textbf{DeSPITE (ours)}& PC & \textbf{69.18} \textit{(+3.24)} \\
        PST-Transformer + \textbf{DeSPIE (ours)} & PC & \textbf{70.26} \textit{(+4.32)}\\
        PST-Transformer + \textbf{DePITE (ours)} & PC & \textbf{70.65} \textit{(+4.71)}\\
        \bottomrule
    \end{tabular}%
    }%
    \caption{HAR classification results on the HMPEAR action recognition dataset, segment-level accuracy Acc(Seg) is reported.}
    \label{tab:hmpear_results}
    
\end{table}

\begin{table}[t]
    \centering
    \resizebox{0.8\columnwidth}{!}{%
    \begin{tabular}{lcc|c}
        \toprule
        \multicolumn{2}{c}{Method} & Modality & Acc(Seg)↑ \\
        \midrule
         \multicolumn{4}{c}{\textit{Uni-Modal Supervised Baselines}} \\
        \midrule
        \multicolumn{2}{c}{$\text{PST-Transformer}^{\dagger}$~\cite{fan2022point}} & PC & 67.38 \\
        \multicolumn{2}{c}{$\text{LSTM}^{\dagger}$}~\cite{hochreiter1997long} & IMU & 65.62 \\
        \multicolumn{2}{c}{$\text{ACTOR}^{\dagger}$}~\cite{petrovich2021action} & Skeleton & 68.23 \\
        \midrule
        \multicolumn{4}{c}{\textit{w/ Zero Shot}} \\
        \midrule
        PST-Transformer &\multirow{3}{*}{\textbf{ + DeSPITE (ours)}}  & PC & 30.42 \\
        LSTM & & IMU & 29.88 \\
        ACTOR & & Skeleton & 34.89 \\
                                                 
        \bottomrule
        \midrule
        \multicolumn{4}{c}{\textit{w/ Linear Probing}} \\
        \midrule
        PST-Transformer & \multirow{3}{*}{\textbf{+ DeSPIE (ours)}} & PC & 67.06 \\
                                         LSTM &   & IMU & 58.29 \\
                                          ACTOR & & Skeleton & 61.76 \\
         
        PST-Transformer & \multirow{3}{*}{\textbf{+ DeSPITE (ours)}} & PC & 67.06 \\
           LSTM &                                          & IMU & 62.06 \\
           ACTOR &                                       & Skeleton & 67.20 \\
        \midrule
         \multicolumn{4}{c}{\textit{w/ Finetuning}} \\
        \midrule
        PST-Transformer & \multirow{3}{*}{\textbf{+ DeSPIE (ours)}} & PC & 67.51 \\
        LSTM &                                         & IMU & \textbf{69.21} \textit{(+3.59)} \\
          ACTOR &                                         & Skeleton & 68.31 \\
         
          PST-Transformer & \multirow{3}{*}{\textbf{+ DeSPITE (ours)}} & PC & \textbf{69.00}  \textit{(+1.62)} \\
                LSTM &                                      & IMU & 68.40 \\
                 ACTOR &                                 & Skeleton & \textbf{70.64} \textit{(+2.41)} \\
        
        \bottomrule
        
    \end{tabular}%
    }%
    \caption{HAR classification results on the Babel-LIPD-v2-CLS action recognition dataset, segment-level accuracy Acc(Seg) is reported.}
    \label{tab:babelcls_results}
    
\end{table}

%\subsubsection{Ablation Probing}

%\subsection{Results: Zero-shot HAR}

\subsection{Qualitative Results: Embedding Space}
\begin{figure}[t]
     \centering
    \begin{subfigure}[b]{0.22\textwidth}
         \centering
         \includegraphics[width=\textwidth]{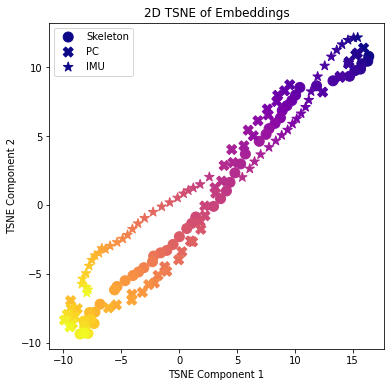}
         \caption{DeSPIE 1 sequence}
         \label{fig:y equals x}
     \end{subfigure}
     \hspace{1em}
     \begin{subfigure}[b]{0.22\textwidth}
         \centering
         \includegraphics[width=\textwidth]{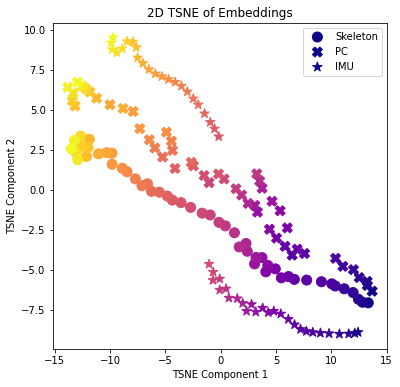}
         \caption{DeSPITE 1 sequence}
         \label{fig:three sin x}
     \end{subfigure}
     \hspace{1em}
     %\vspace{-1cm}
     %\begin{subfigure}[b]{0.22\textwidth}
     %    \centering
      %   \includegraphics[width=\textwidth]{ICCV2025-Author-Kit/figures/embeddings/spie_multi.png}
     %    \caption{DeSPIE 10}
     %    \label{fig:y equals x}
     %\end{subfigure}
     %\hspace{1em}
     %\begin{subfigure}[b]{0.22\textwidth}
     %    \centering
     %    \includegraphics[width=\textwidth]{ICCV2025-Author-%Kit/figures/embeddings/spite_multi.png}
      %   \caption{DeSPITE 10}
      %   \label{fig:three sin x}
     %\end{subfigure}
    %\hspace{1em}
    \begin{subfigure}[b]{0.22\textwidth}
         \centering
         \includegraphics[width=\textwidth]{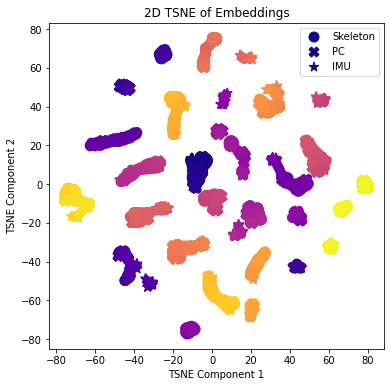}
         \caption{DeSPIE 20 sequences}
         \label{fig:y equals x}
     \end{subfigure}
     \hspace{1em}
     \begin{subfigure}[b]{0.22\textwidth}
         \centering
         \includegraphics[width=\textwidth]{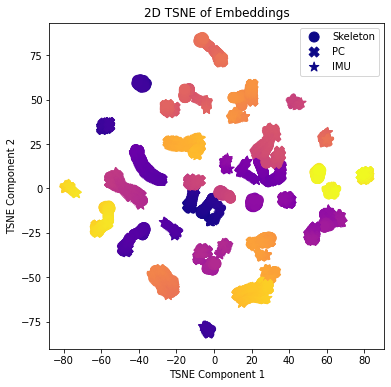}
         \caption{DeSPITE 20 sequences}
         \label{fig:three sin x}
     \end{subfigure}

     \caption{Using TSNE, we visualize the joint embedding space between skeletons ($\bullet$), point clouds ($\times$), and IMU ($\star$) on 1 (a,b) and 20 (c,d) randomly sampled sequences with 50 consecutive sequences of both DeSPIE (left) and DeSPITE (right). Each point is colored by its time index with a colormap to emphasize the similarity among each of the modalities over time. }
     \label{fig:embeddings_2}
\end{figure}

Using TSNE~\cite{van2008visualizing}, we analyze the learned embedding space of DeSPITE and DeSPIE. 
%to understand modality alignment and the impact of text embeddings on association. 
Figure~\ref{fig:embeddings_2} (a, b) shows embeddings of the same 50-frame sequence per modality (skeletons $\bullet$, point clouds $\times$, IMU $\star$), where both models exhibit strong cross-modal alignment, although DeSPIE demonstrates tighter associations. Figure~\ref{fig:embeddings_2} (c, d) extends this to 20 sequences, revealing distinct clusters that indicate semantic motion encoding. However, DeSPIE’s embeddings are more distinct, qualitatively supporting our retrieval findings and confirming that text embeddings negatively affect matching and temporal moment retrieval performance.

\subsection{Qualitative Results: Retrieval from AMASS and LIPD Database} \label{sec:exp_retrieval}
\begin{figure}[t]
    \centering
    \includegraphics[width=\linewidth]{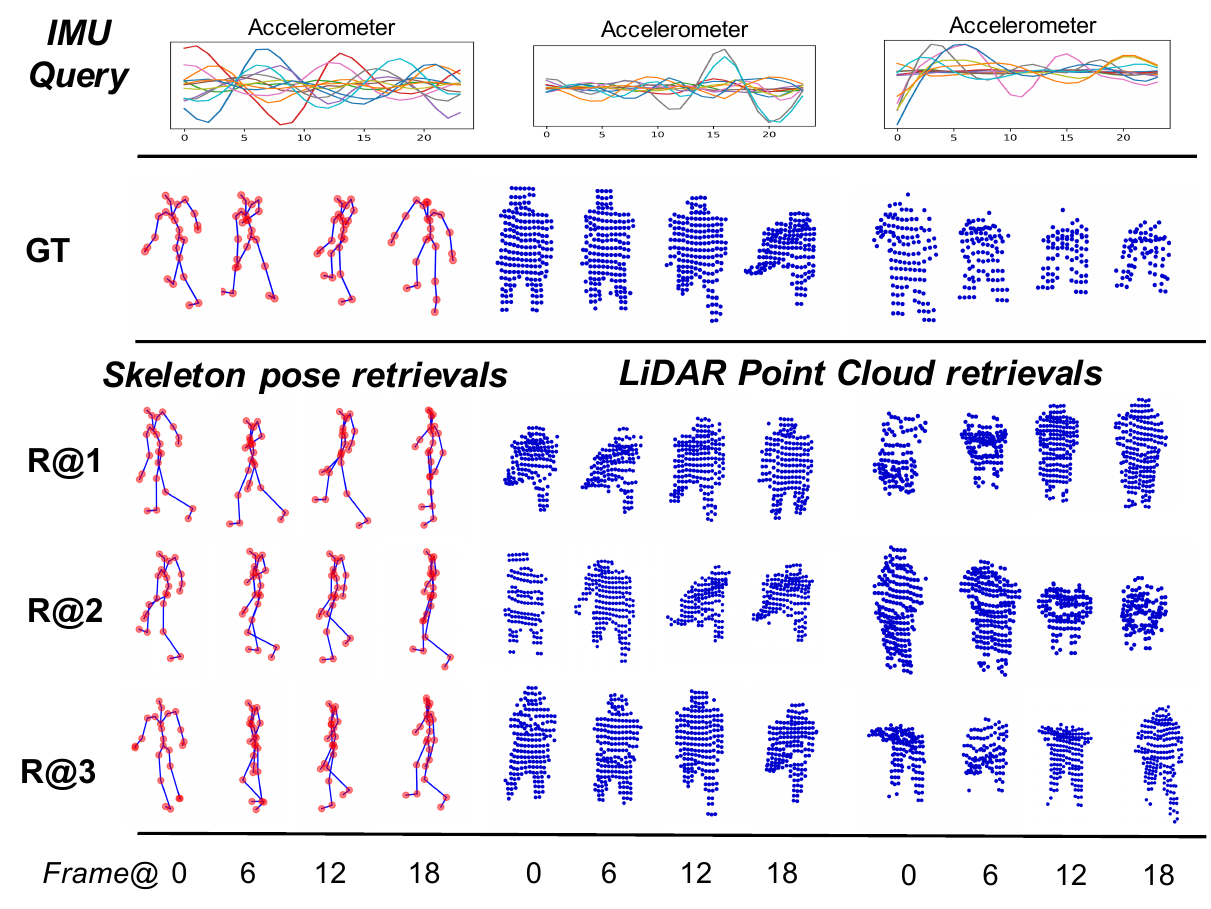}
    \caption{IMU$\to$Skeleton and IMU$\to$Point cloud Retrieval from AMASS and LIPD database, respectively.}
    \label{fig:imu_retrieval}
\end{figure}

%Our method can effectively query the poses in the full AMASS dataset to retrieve skeleton representations that correspond to IMU time series or point cloud sequences. 
Figure~\ref{fig:imu_retrieval} shows that we can interpret IMU signals using our method by querying a large motion database like AMASS or LIPD. We show retrievals of skeletons from AMASS and point clouds from LIPD using IMU (top) as a query, also showing the ground truth. The retrievals semantically capture the motion performed by the IMU signal, allowing us to understand that the IMU signal corresponds to walking while turning (left), doing a lunge (middle), and forward stretch (right), respectively. Extending IMU2CLIP~\cite{moon2023imu2clip}, our method can help interpret IMU signals with skeletons and point cloud sequences. 

%The retrievals represent the overall motion of a moving pedestrian that turns to the left. %This experiments shows the potential for future work and real-world applications, where the IMU signal is very hard to interpret, but the motion representations in point cloud or skeletons easily reveal the activity of the person. The retrievals capture the overall motion, which consists of a moving pedestrian that turns to the left.

\begin{figure}[t]
    \centering
    \includegraphics[width=\linewidth]{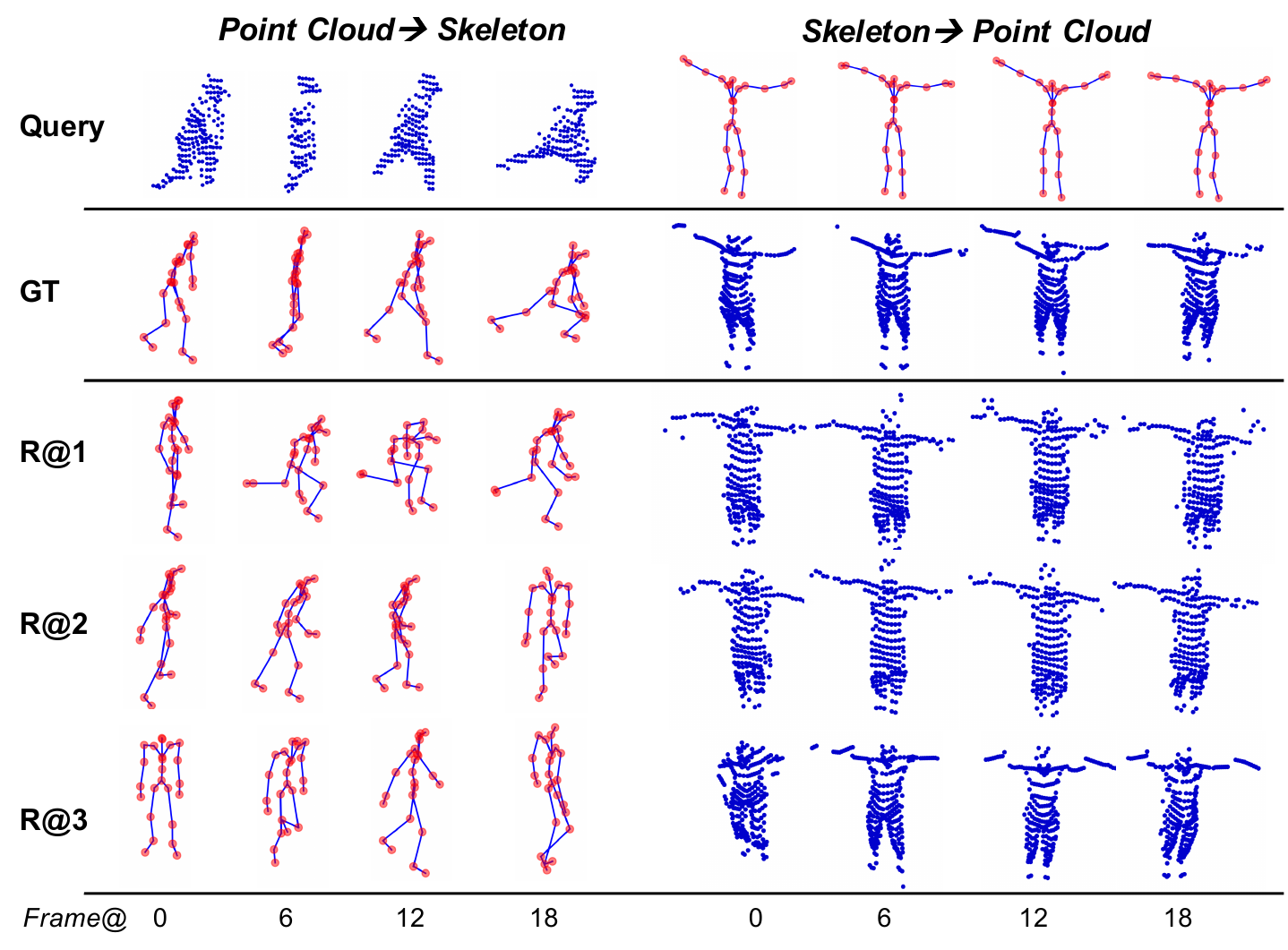}
    \caption{Point cloud$\to$Skeleton and Skeleton$\to$Point cloud Retrieval from AMASS and LIPD database, respectively.}
    \label{fig:pointcloud_skeleton_retrieval}
\end{figure}

Figure~\ref{fig:pointcloud_skeleton_retrieval} shows the corresponding retrievals between skeleton and point clouds. On the left,
a pedestrian performs a ``lunge'' motion, captured effectively by the retrieved skeletons from the AMASS database. On the right, a pedestrian performs a t-pose and then moves his arm into a normal standing position. The retrieved point clouds from the LIPD database follow these motions, showing a learned correspondence of motion between both modalities. We can see that different motion sequences are retrieved with different point cloud densities. \textbf{\textit{To better assess the effectiveness of DeSPITE for cross-modal retrieval}}, we provide animated videos in the supplementary material.
%we provide 50 randomly sampled retrievals (50 for each IMU$\to$Point cloud, IMU$\to$Skeleton, Point cloud$\to$Skeleton, Skeleton$\to$Point cloud) for each modality as animated videos in the attached supplementary material. 

\iffalse
%Using LIPD-Babel-v1, we use the whole training set as a database, and qualitatively show topk retrievals using embeddings from the test set. 
\begin{figure*}[t]
    \centering
    \includegraphics[width=\linewidth]{figures/retrieval/placeholder.png}
    \caption{Top 10 retrievals from the database using embeddings of the unseen test set. We show the corresponding retrievals w.r.t. IMU$\rightarrow$Skeleton, Skeleton$\rightarrow$Skeleton, PCD$\rightarrow$Skeleton}
    \label{fig:retrieval}
\end{figure*}
\fi

\subsection{Qualitative Results: Temporal Moment Retrieval}

\begin{figure}[t]
    \centering
    \includegraphics[width=\linewidth]{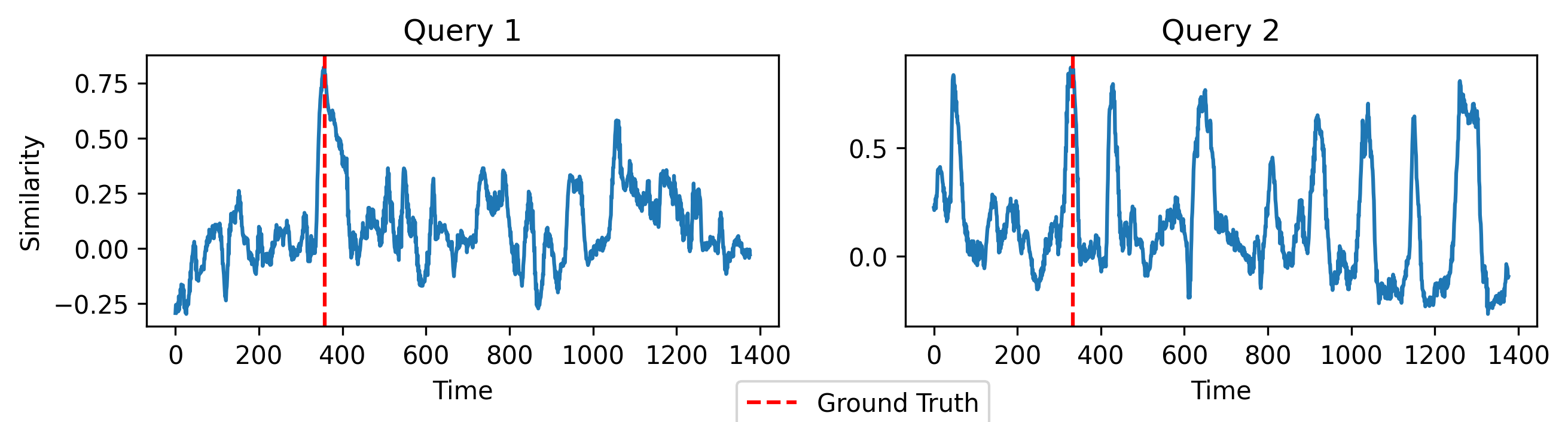}
    \caption{We show two random IMU queries to localize the respective moment in a 1400 frame-long point cloud sequence from the unseen LIPD test.}
    \label{fig:qualitative_temporal_moment}
\end{figure}

Figure~\ref{fig:qualitative_temporal_moment} illustrates how effectively an encoded IMU query can retrieve relevant moments in a point cloud video. We visualize cosine similarity across a 1400-frame sequence containing diverse activities, with peaks aligning precisely with the ground truth timestamps. Despite no explicit training for this, in Query 2, our approach identifies repeated instances of a person standing still, highlighting the ability of DeSPITE to encode semantic meaning for certain activities within the embedding space. \textbf{\textit{We provide an animated video}} of this in the supplementary material (``temporal\_moment\_retrieval/temporal\_retrieval\_imu.gif''). %showing the retrieved moments at the similarity peaks.

\section{Conclusion}

We introduce DeSPITE, a \underline{\textbf{De}e}p \underline{\textbf{S}}keleton-\underline{\textbf{P}}ointcloud-\underline{\textbf{I}}MU-\underline{\textbf{T}}ext \underline{\textbf{E}}mbedding model, enabling novel cross-modal tasks such as matching for re-identification, temporal moment retrieval, and retrieval across modalities. Unlike prior RGB-centric approaches, DeSPITE leverages LiDAR point clouds as the primary visual modality for contrastive learning. On the constructed LIPD-Babel dataset, we establish strong baselines for these tasks, providing a foundation for future comparisons. Additionally, we demonstrate that contrastive pre-training for HAR with DeSPITE achieves new state-of-the-art performance on MSR-Action3D and HMPEAR. Our findings highlight that text-enhanced embeddings benefit HAR but slightly limit retrieval performance compared to text-free variants. This work paves the way for general-purpose LiDAR-based video encoders for human activity understanding.

%We introduce DeSPITE, a \underline{\textbf{D}}eep \underline{\textbf{S}}keleton-\underline{\textbf{P}}ointcloud-\underline{\textbf{I}}MU-\underline{\textbf{T}}ext \underline{\textbf{E}}mbedding model, enabling novel tasks not possible before between these modalities, such as matching for re-identification, temporal moment retrieval, and cross-modal retrieval. Unlike prior works that rely on RGB as the main visual modality, DeSPITE explores LiDAR point clouds as the main modality for cross-modal contrastive learning. Through LIPD-Babel, we evaluated DeSPITE in all aforementioned novel tasks, enabling future work to compare against our baselines. Furthermore, we show that contrastive pre-training for HAR with DeSPITE yields new state-of-the-art performance in MSR-Action3D and HMPEAR, showing that learned representations improve performance across multiple settings. Our main findings in this paper are that a joint embedding space with text helps for downstream HAR tasks, but slightly limits cross-modal retrieval and matching tasks compared to variants without text. This paper paves the way for future work on general-purpose pre-trained LiDAR point cloud video encoders for human activity understanding.

{
    \small
    \bibliographystyle{ieeenat_fullname}
    \bibliography{main}
}

\clearpage
\setcounter{page}{1}
\maketitlesupplementary

\section{More Information on LIPD-Babel}\label{app:dataset}
\begin{table}[t]
    \centering
    \begin{tabular}{c|cc}
    \toprule
        & LIPD-Babel-v1 & LIPD-Babel-v2 \\
        \midrule
        \#Sequences Train & 502958 & 403430 \\
        \#Sequences Test & 85551 & 58802 \\
        \midrule
        \#Text Train & 187641 & 135699 \\
        \#Text Test & - & 58802 \\
        \bottomrule
        %\#Classes Train &  &  \\
        %\#Classes Test & - & 58802 \\
    \end{tabular}
    \caption{Number of total training sequences and frame-wise text labels when considering 24 frame sequences at 10 fps in LIPD-Babel-v1 and LIPD-Babel-v2.}
    \label{tab:statistics_dataset}
\end{table}

LIPD~\cite{ren2023lidar} is a large-scale dataset combining LiDAR point clouds, IMU, and skeleton poses. It includes a mix of real and synthetic LiDAR point clouds and IMU measurements, taking advantage of the AMASS~\cite{AMASS:ICCV:2019} motion capture dataset. It combines their own recorded real sequences with ground truth SMPL~\cite{loper2023smpl} pose parameters from DIP-IMU~\cite{huang2018deep}, LiDARHuman26M~\cite{ren2023lidar}, AIST++\cite{li2021ai}, and a subset of AMASS~\cite{AMASS:ICCV:2019} (including ACCAD, BML-Movi, CMU, and TotalCapture(TC)~\cite{trumble2017total}). From this large collection of data, DIP-IMU, TC, the testing set of LiDARHuman26M, and their LIPD test set are used for evaluation, while the remaining data is used as the training set. From the SMPL poses of the AMASS subsets and AIST++, LIPD has generated synthetic point clouds and IMU recordings. For DIP-IMU, they generated synthetic point clouds, and took the real IMU recordings. For LiDARHuman26M, real point clouds are provided, and LIPD generated synthetic IMU recordings. LIPD has provided more details on the generation in the supplementary material of their work but did not publish the code. 

While LIPD is a large-scale data for human motion with LiDAR, IMU, and skeletal poses, it is missing activity annotations. Fortunately, the Babel dataset~\cite{BABEL:CVPR:2021} has annotated the AMASS dataset with strong efforts with frame-wise activity annotations. Taking advantage of these annotations, we have merged the Babel annotations into the corresponding AMASS subsets present in LIPD, i.e., ACCAD, BML-Movi, CMU, and TC. 

To achieve this, we take advantage of the specific sequence ID's for each AMASS sequence that are stored both in LIPD and in Babel. This allows a unique mapping between both datasets, allowing to add text annotations to the AMASS subset in LIPD. More specifically, all AMASS sequence ids follow the pattern of ``dataset/sequencecategory/posesequence\_poses.npz''. For example, ''ACCAD/Female1Gestures\_c3d/D2\_-\_Wait\_1\_poses.npz'' or ``CMU/118/118\_17\_poses.npz''. In LIPD, these sequence ids were modified to follow the pattern ``dataset/sequencecategory/posesequence\_stageii'', leading to ''ACCAD/Female1Gestures\_c3d/D2\_-\_Wait\_1\_stageii'' ``CMU/118/118\_17\_stageii''. Therefore, to achieve the mapping, a reformatting is required by replacing ``stageii'' with ``poses.npz''. Babel follows the exact same format as AMASS, allowing a straightforward mapping form the AMASS subset in LIPD to the respective subset in Babel.

The next difference is the sampling rate of the dataset. AMASS is provided at a higher FPS up to 120FPS, which is much higher than the 10 FPS of LIPD, while Babel is annotated at 30FPS. As a result, we downsample the Babel annotations accordingly to 10FPS to align them with the LIPD dataset. To verify that both datasets are actually temporally aligned after downsampling Babel to 10 FPS, we carefully verified qualitatively that the skeleton poses of the downsampled Babel versions correspond to the poses in the LIPD dataset by plotting them next to each other, and inspecting several sequences from each dataset manually. 

When combning Babel and LIPD to LIPD-Babel, we obtain two versions of the dataset. First, LIPD-Babel-v1, which follows the exact test split of LIPD, and LIPD-Babel-v2, which follows the train/val/test split of Babel. LIPD-Babel-v1 allows to evaluate the respective matching and retrieval tasks, while LIPD-Babel-v2 allows to evaluate classification tasks with labels for both training and testing set. More specifically, for LIPD-Babel-v2, the official training split of Babel is used for the training set, and the validation split for the testing set. The annotations for the test split are not publicly available, which is why we use the validation set as the testing set replacement.

Finally, we preprocess the whole dataset into 24-frame sliding window subsequences to ease the training and testing process. A summary of the number of sequences and corresponding text annotations are provided in Table~\ref{tab:statistics_dataset}. LIPD-Babel-v1 has slightly more sequences than LIPD-Babel-v2 because the test set of Babel is not publicly available, because of which we remove the sequences from ACCAD, BML-movi, CMU, and TC that do not have annotations.

\section{Specific Performance Scores for Matching and Temporal Moment Retrieval} \label{app:scores}
\begin{figure*}[t]
    \centering
    \includegraphics[width=\linewidth]{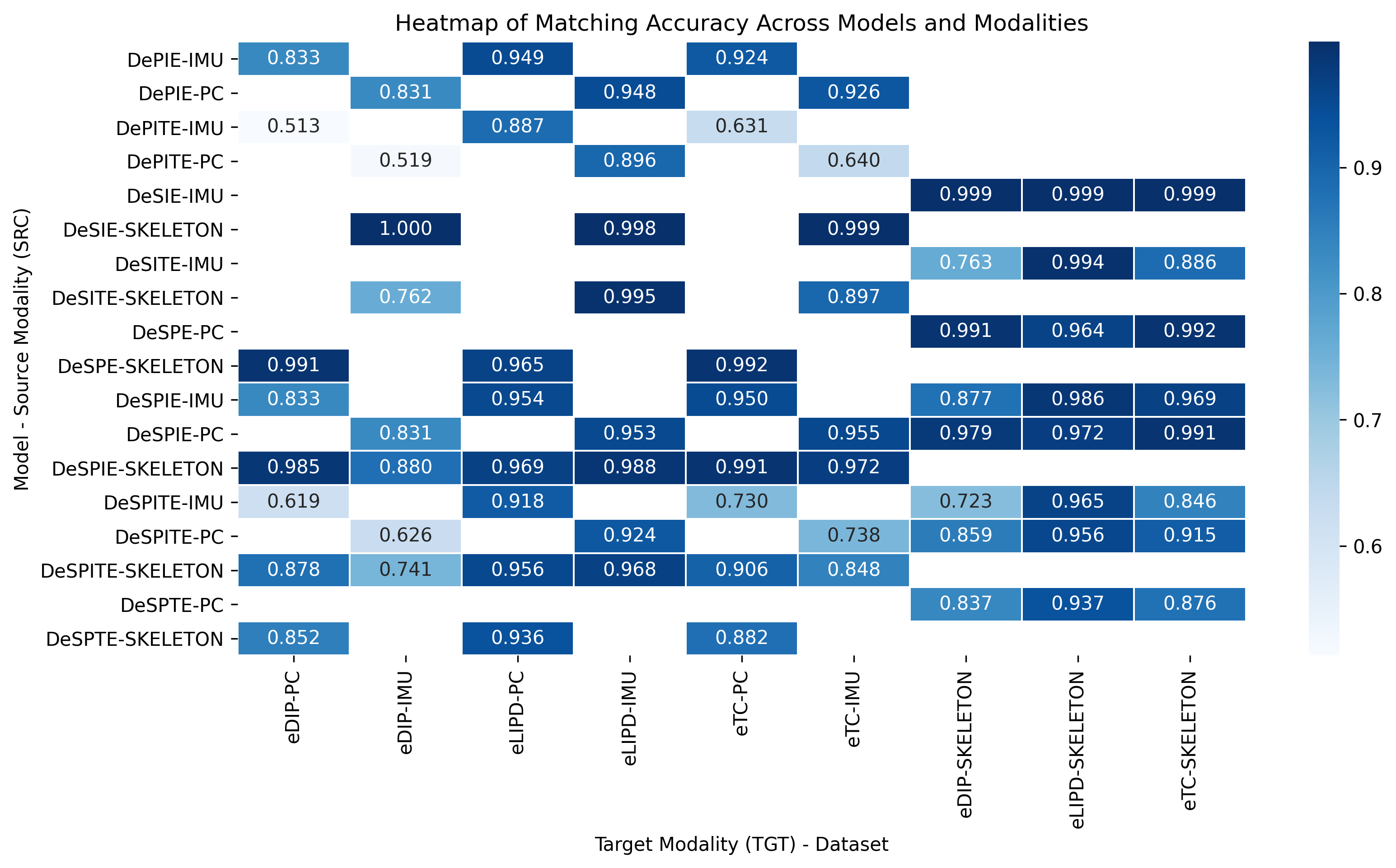}
    \caption{Heatmap to visualize the respective matching results on average across all modalities and datasets at a glance}
    \label{fig:matching_scores}
\end{figure*}
Figure~\ref{fig:matching_performance} and Figure~\ref{fig:moment_retrieval} in the main paper effectively visualizes the differences in the performance of each model and parameters for matching and temporal moment retrieval, which makes a comparison at the scale of the amount of different parameters that we have evaluated easier to see. 

\begin{figure*}[t]
    \centering
    \includegraphics[width=\linewidth]{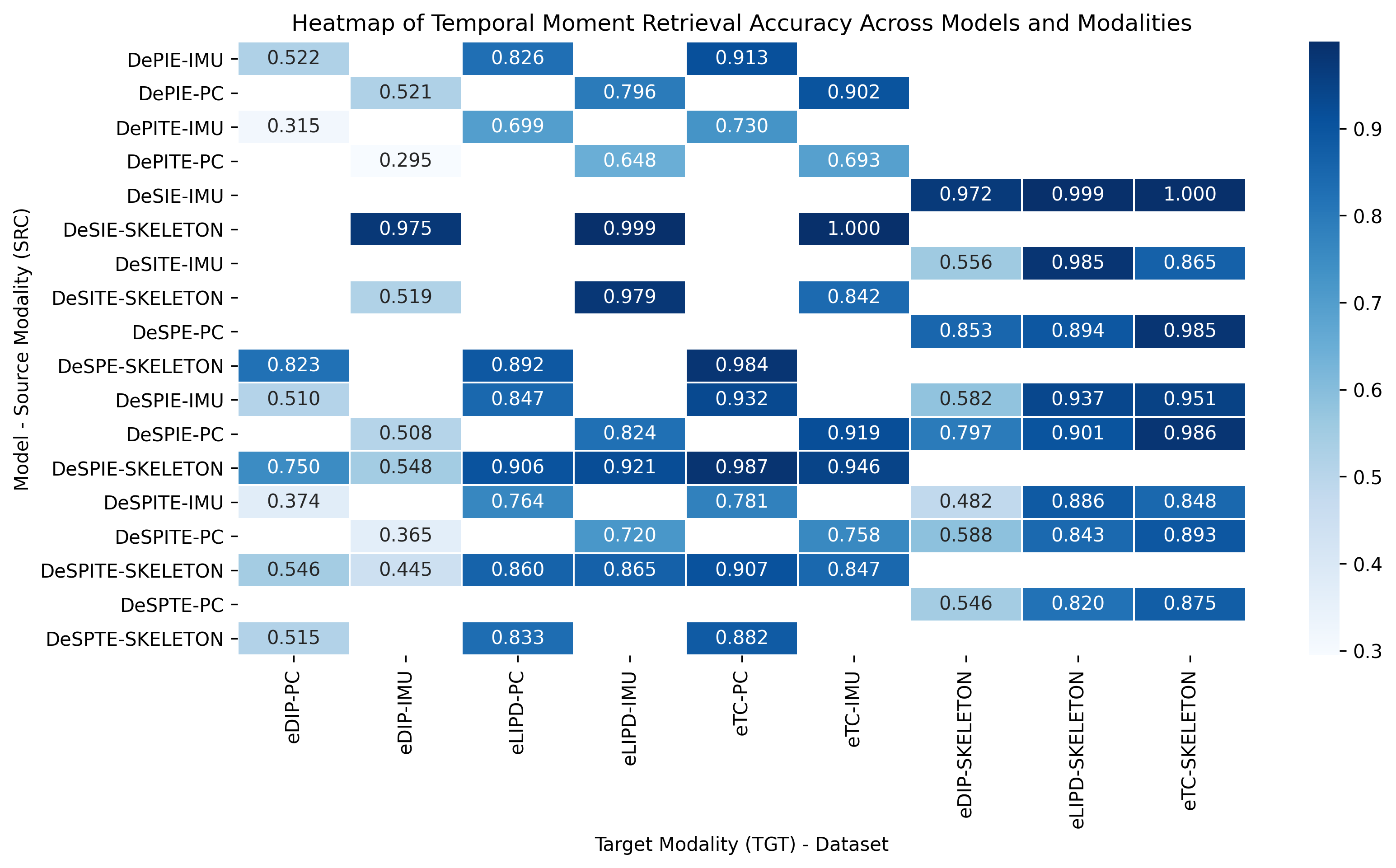}
    \caption{Heatmap to visualize the respective temporal moment retrieval results on average across all modalities and datasets at a glance}
    \label{fig:temporal_scores}
\end{figure*}

In this section of our supplementary material, we provide the specific matching scores between each modality and each dataset through a heatmap in Figure~\ref{fig:matching_scores} to provide quantitative numbers for future work to compare against our baselines. In the same way, we present the specific temporal moment retrieval scores between each modality and each dataset through a heatmap in Figure~\ref{fig:temporal_scores}. These results are the average over the number of subjects for matching and k-shots for temporal moment retrieval. All individual results for each number of subjects for matching, and top-k for temporal moment retrieval are provided in Figures 11-18, and Figures 19-22, respectively.

\begin{figure}
    \centering
    \includegraphics[width=\linewidth]{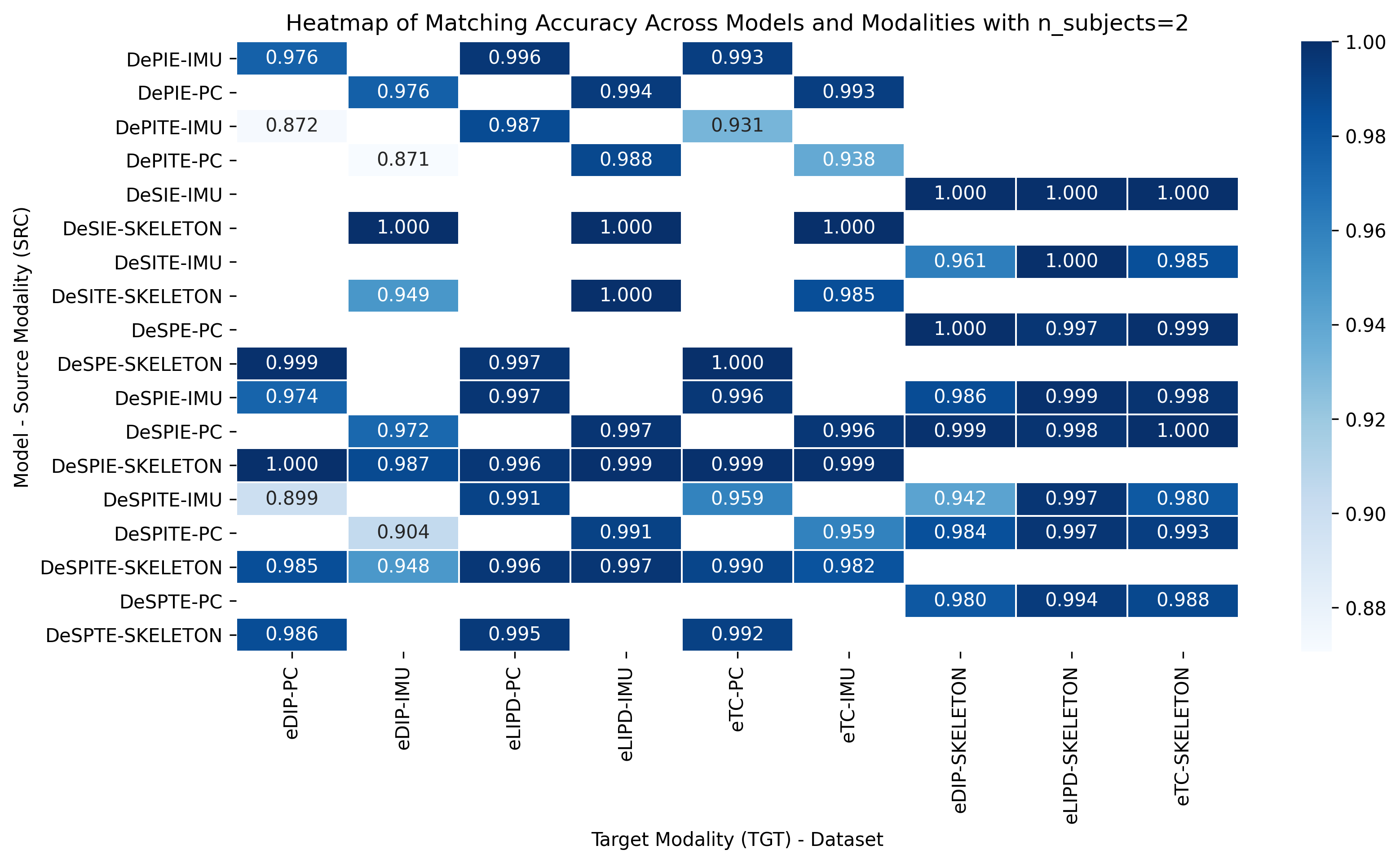}
    \caption{Matching, subjects=2}
    \label{fig:enter-label}
\end{figure}

\begin{figure}
    \centering
    \includegraphics[width=\linewidth]{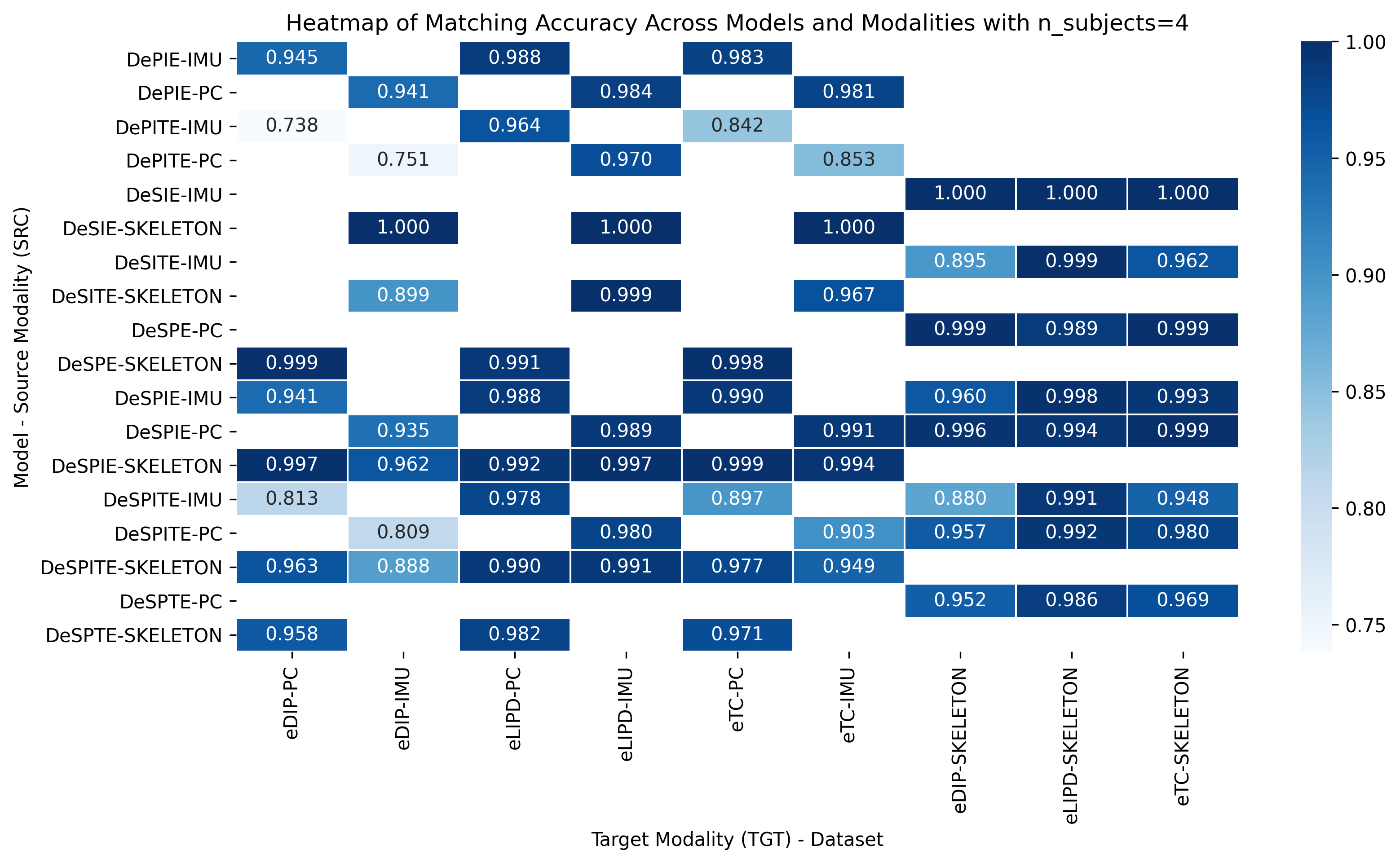}
    \caption{Matching, subjects=4}
    \label{fig:enter-label}
\end{figure}

\begin{figure}
    \centering
    \includegraphics[width=\linewidth]{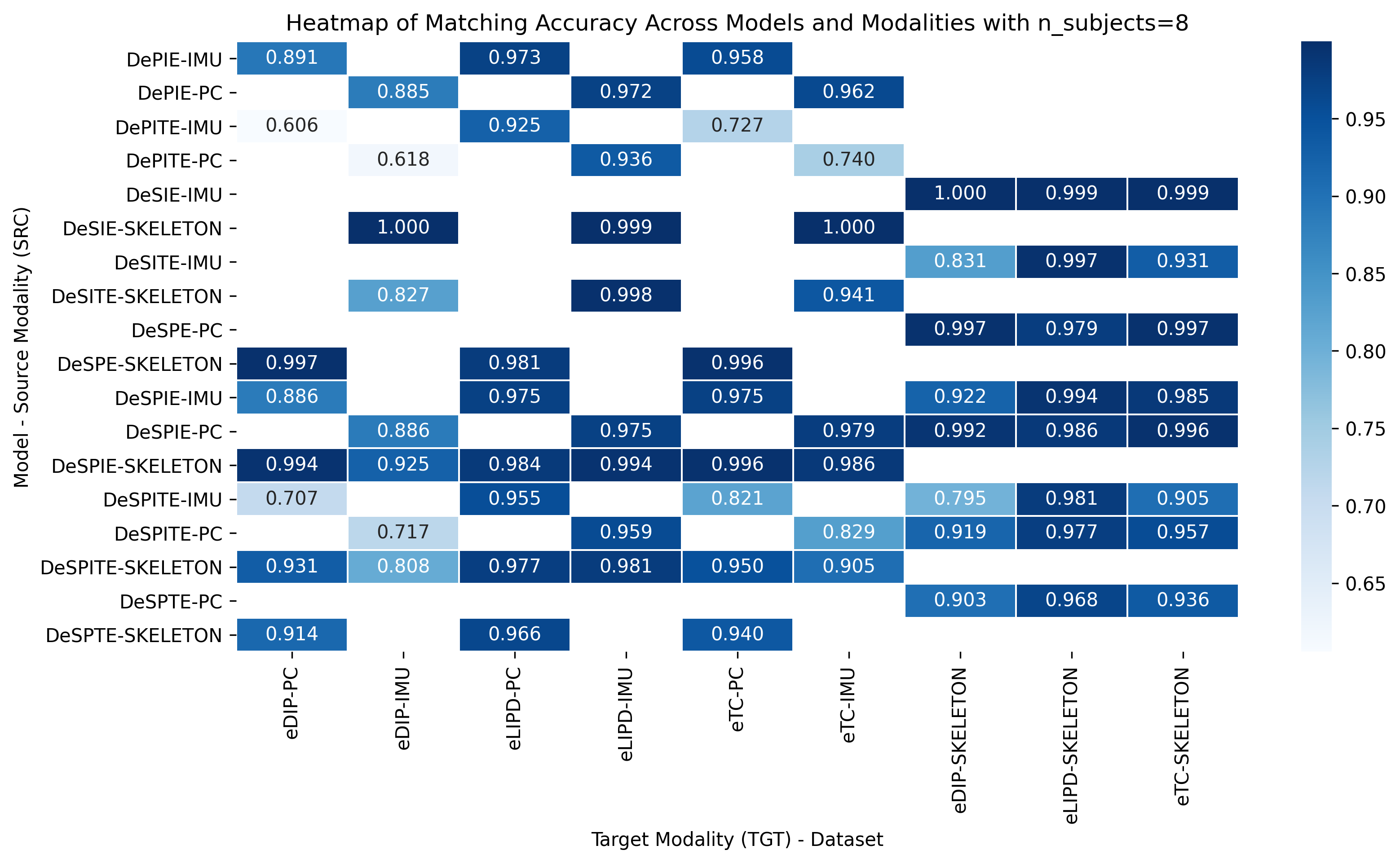}
    \caption{Matching, subjects=8}
    \label{fig:enter-label}
\end{figure}

\begin{figure}
    \centering
    \includegraphics[width=\linewidth]{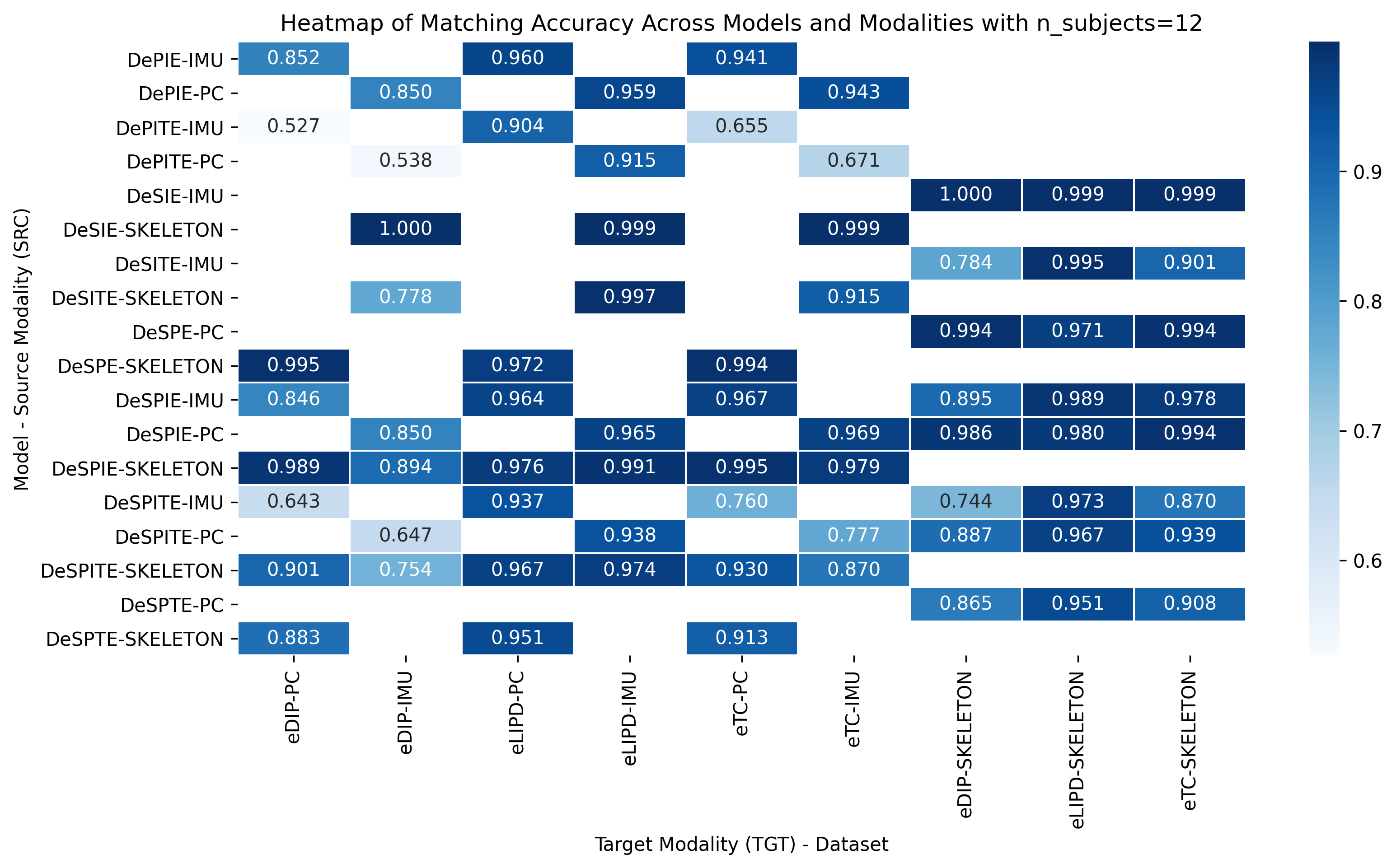}
    \caption{Matching, subjects=12}
    \label{fig:enter-label}
\end{figure}

\begin{figure}
    \centering
    \includegraphics[width=\linewidth]{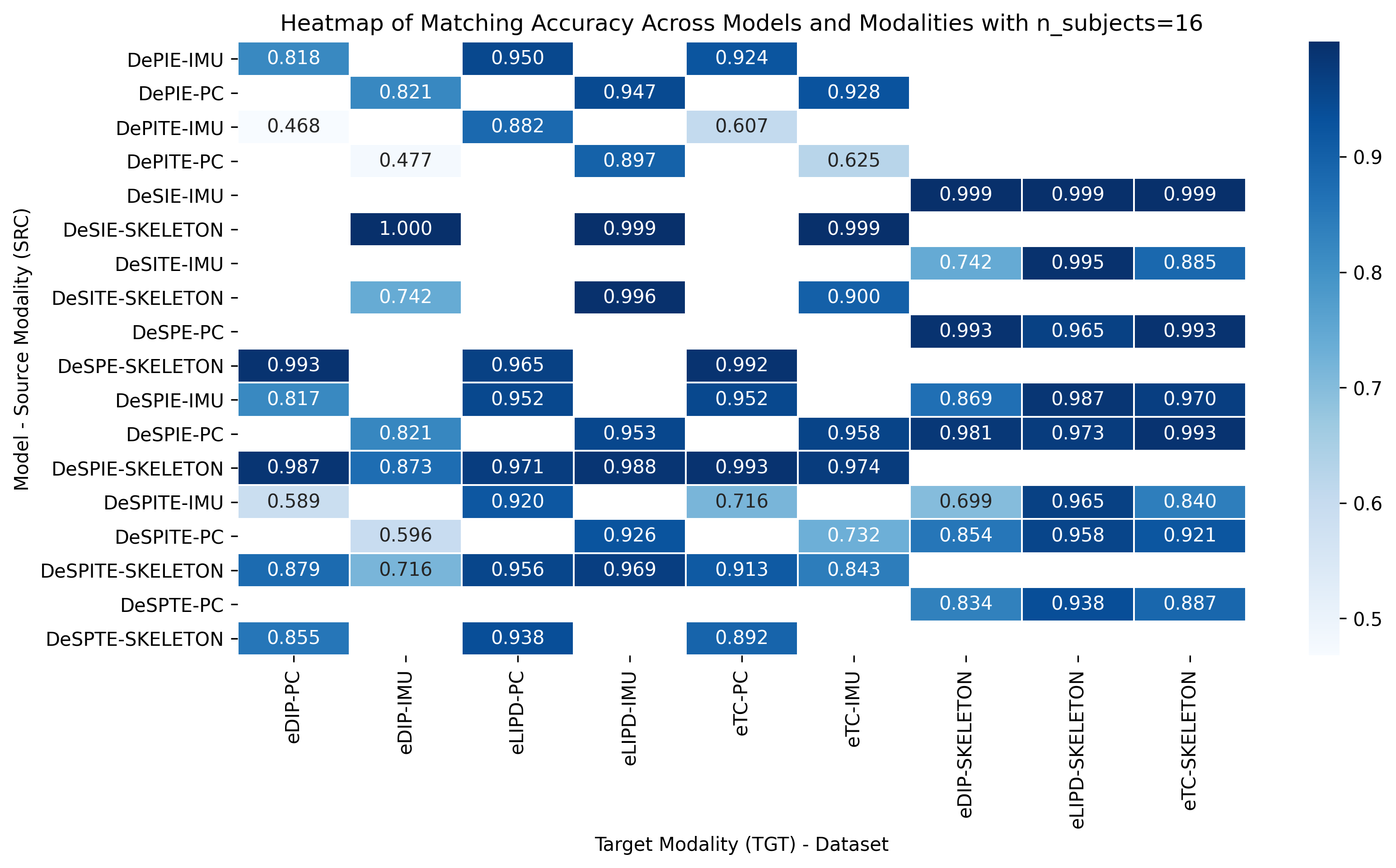}
    \caption{Matching, subjects=16}
    \label{fig:enter-label}
\end{figure}

\begin{figure}
    \centering
    \includegraphics[width=\linewidth]{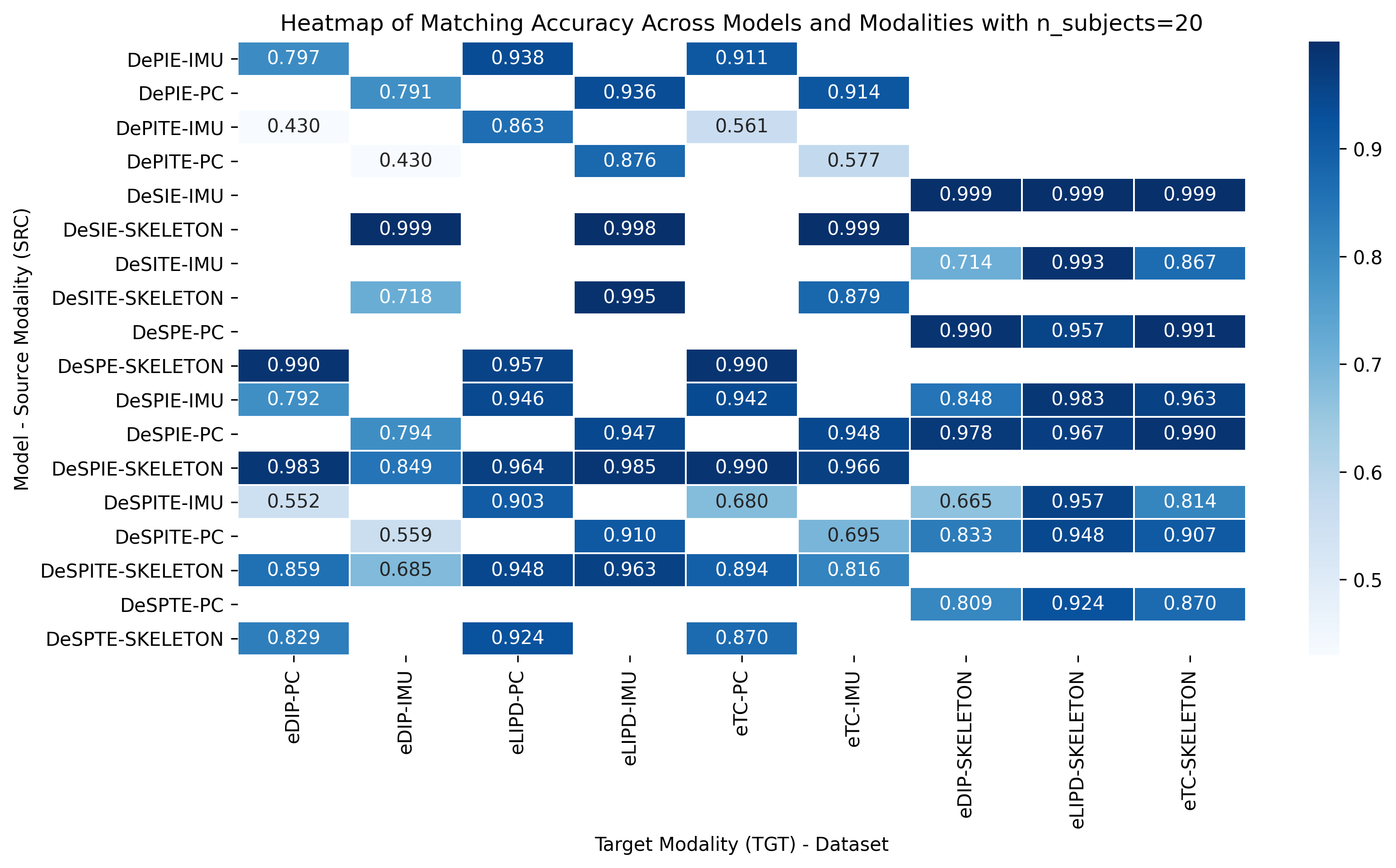}
    \caption{Matching, subjects=20}
    \label{fig:enter-label}
\end{figure}

\begin{figure}
    \centering
    \includegraphics[width=\linewidth]{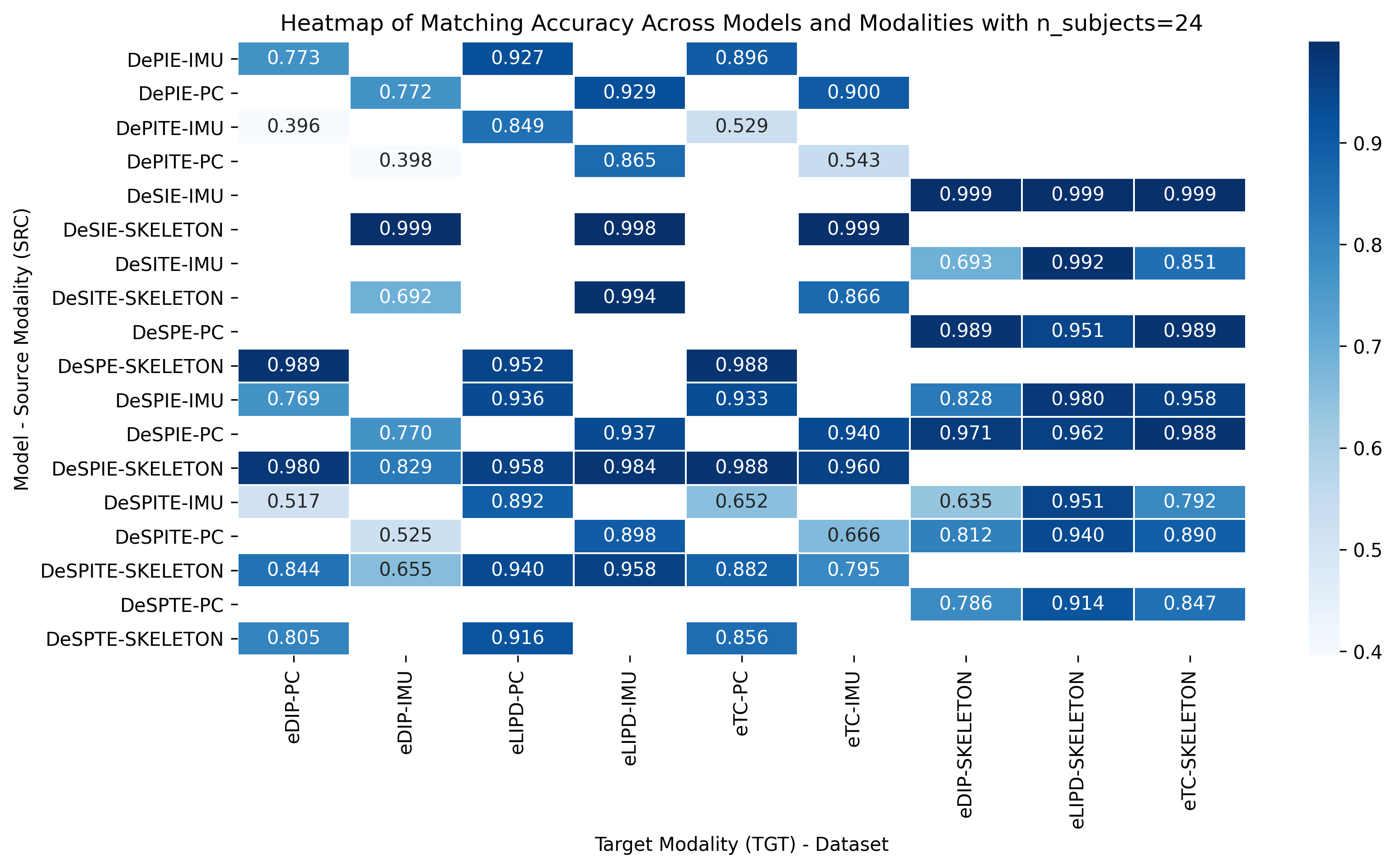}
    \caption{Matching, subjects=24}
    \label{fig:enter-label}
\end{figure}

\begin{figure}
    \centering
    \includegraphics[width=\linewidth]{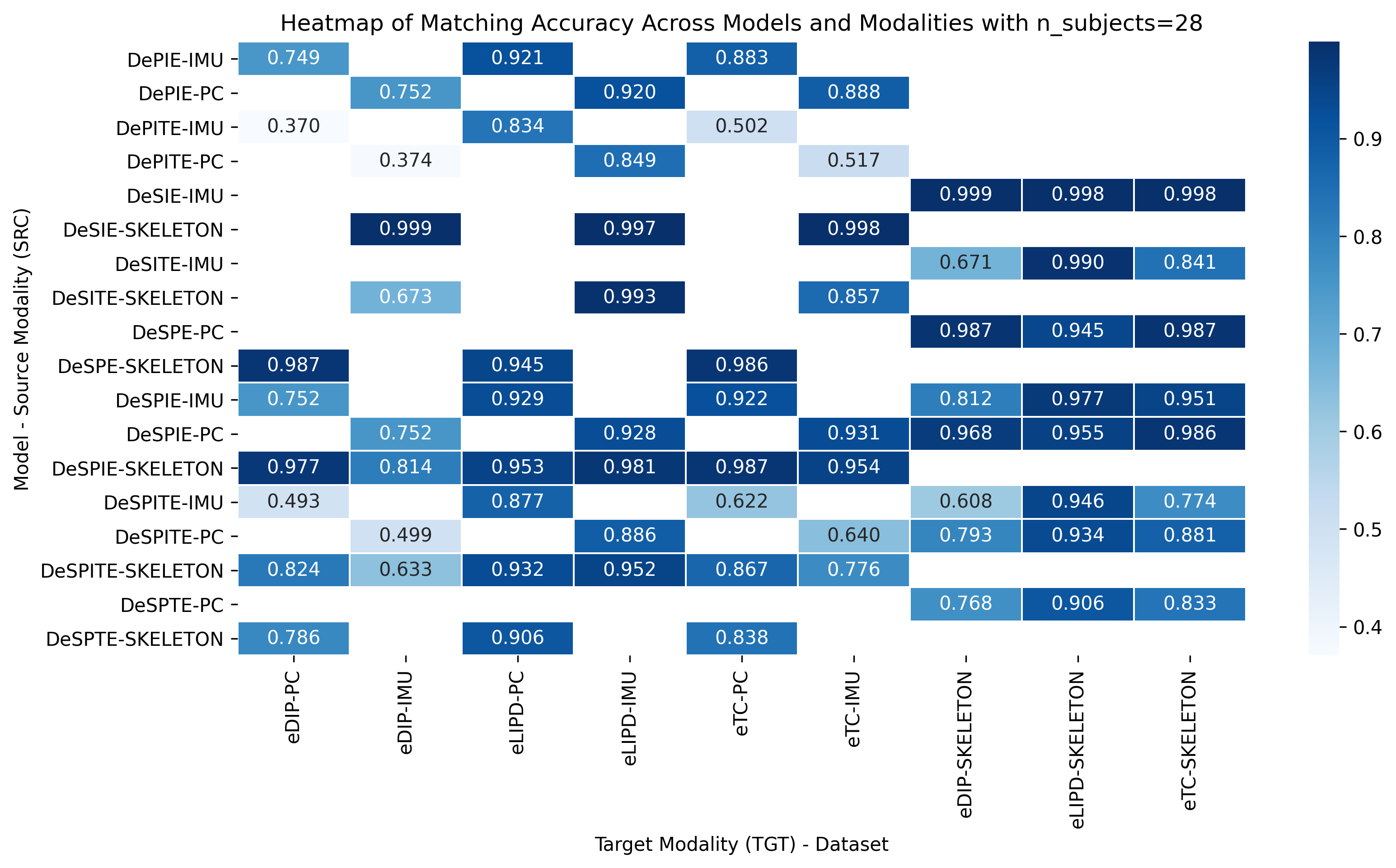}
    \caption{Matching, subjects=28}
    \label{fig:enter-label}
\end{figure}

\begin{figure}
    \centering
    \includegraphics[width=\linewidth]{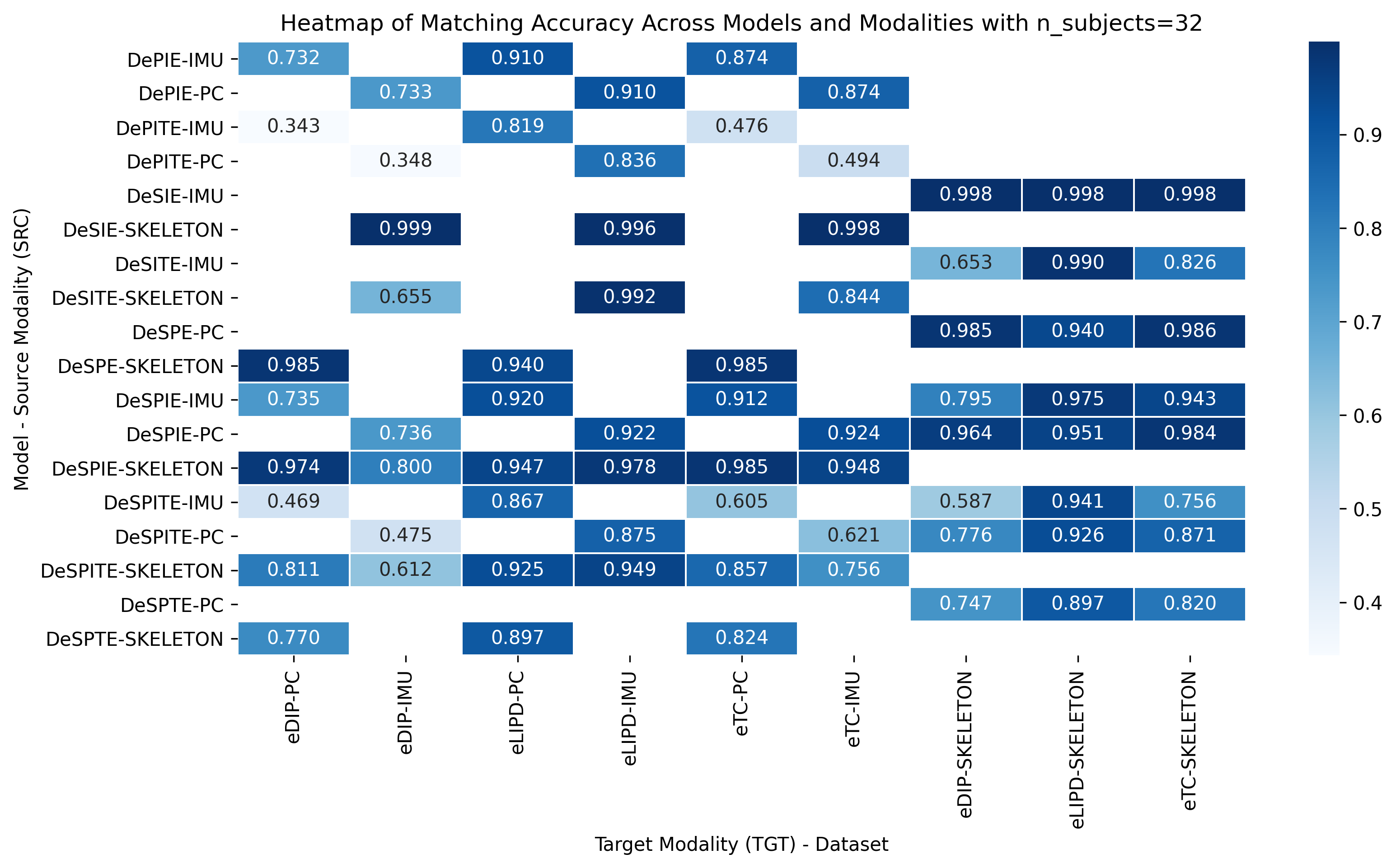}
    \caption{Matching, subjects=32}
    \label{fig:enter-label}
\end{figure}

%%%%%%%% TEMPORAL MOMENT RETRIEVAL

\begin{figure}
    \centering
    \includegraphics[width=\linewidth]{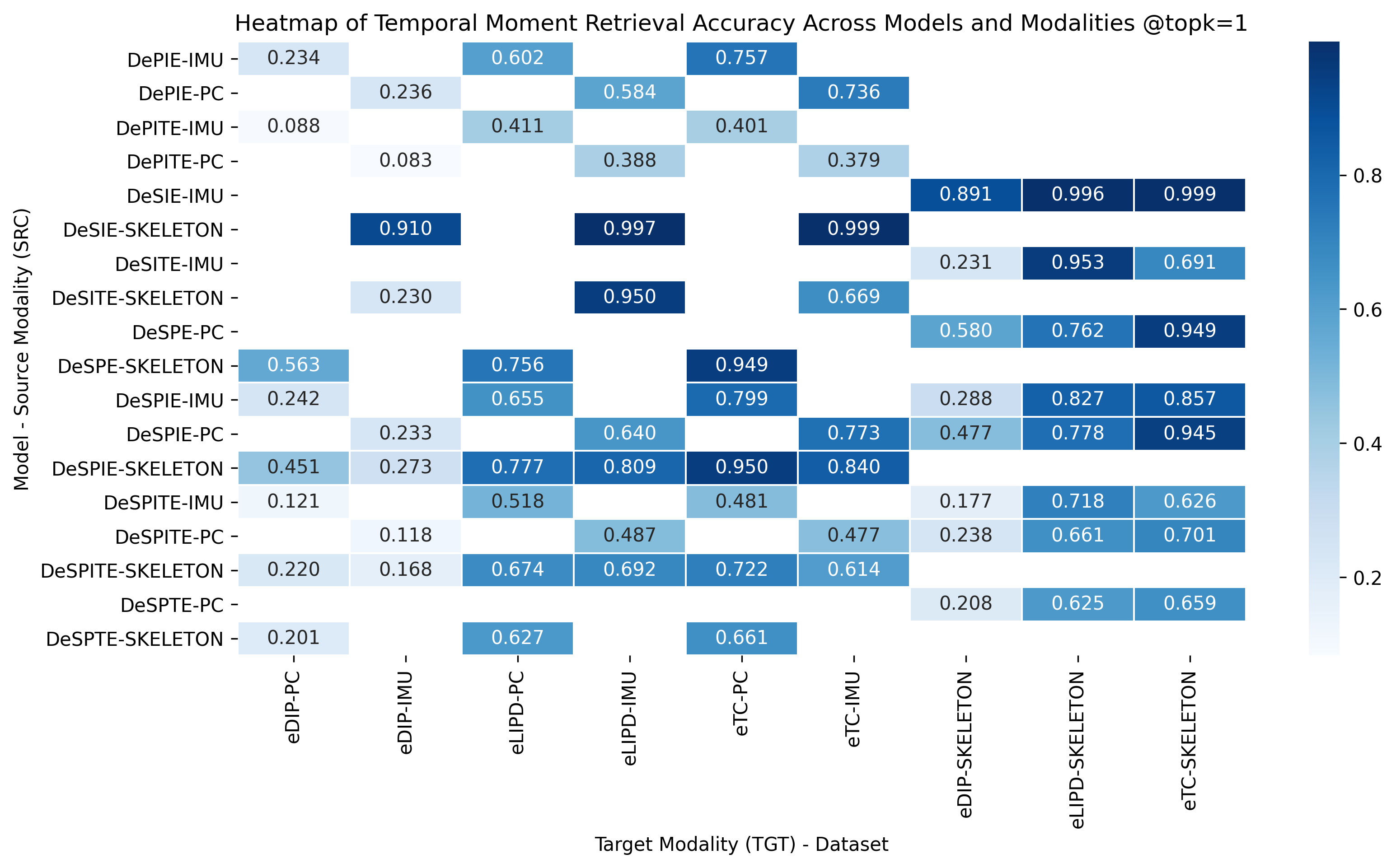}
    \caption{Temporal Moment Retrieval, topk=1}
    \label{fig:enter-label}
\end{figure}

\begin{figure}
    \centering
    \includegraphics[width=\linewidth]{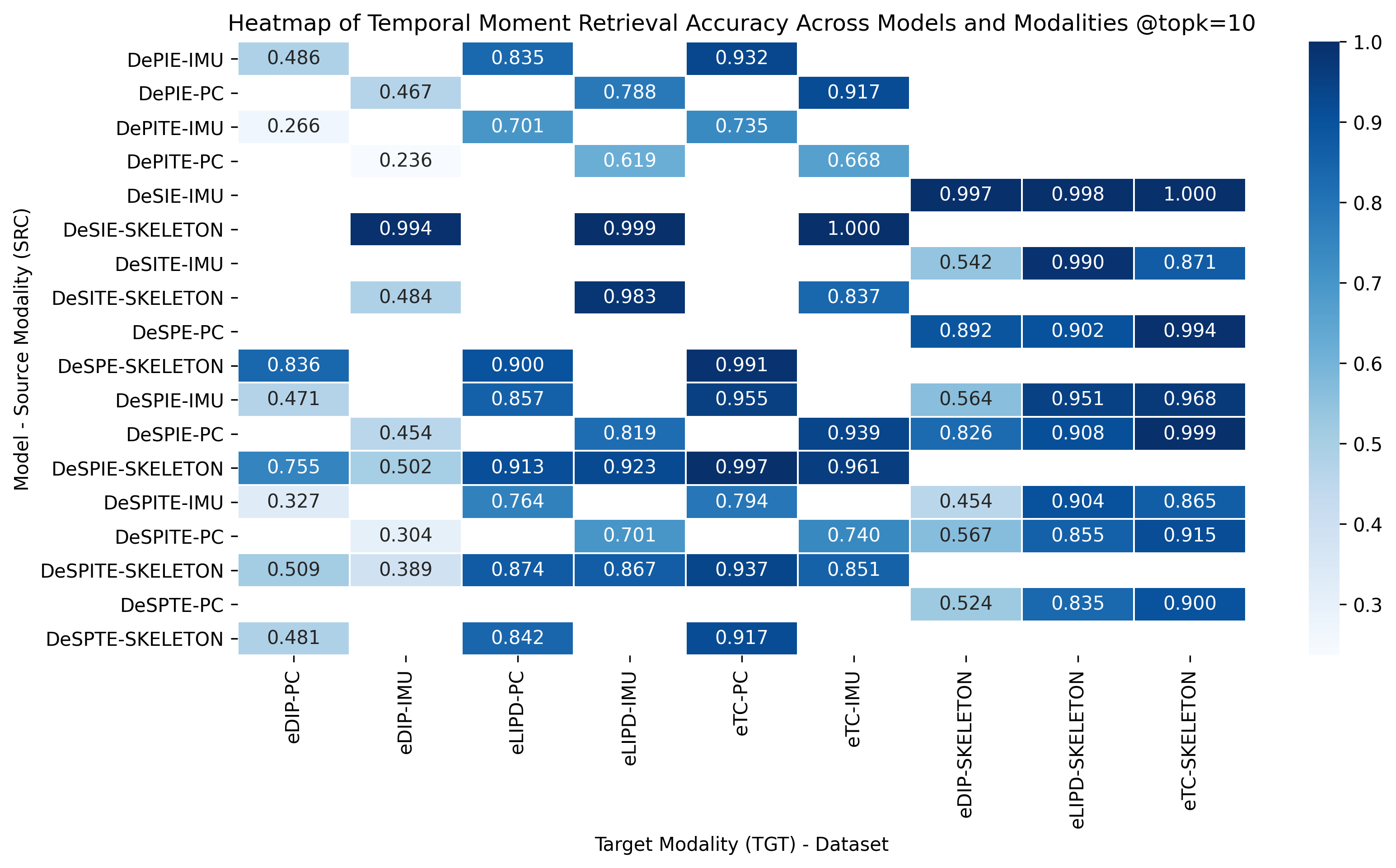}
    \caption{Temporal Moment Retrieval, topk=10}
    \label{fig:enter-label}
\end{figure}

\begin{figure}
    \centering
    \includegraphics[width=\linewidth]{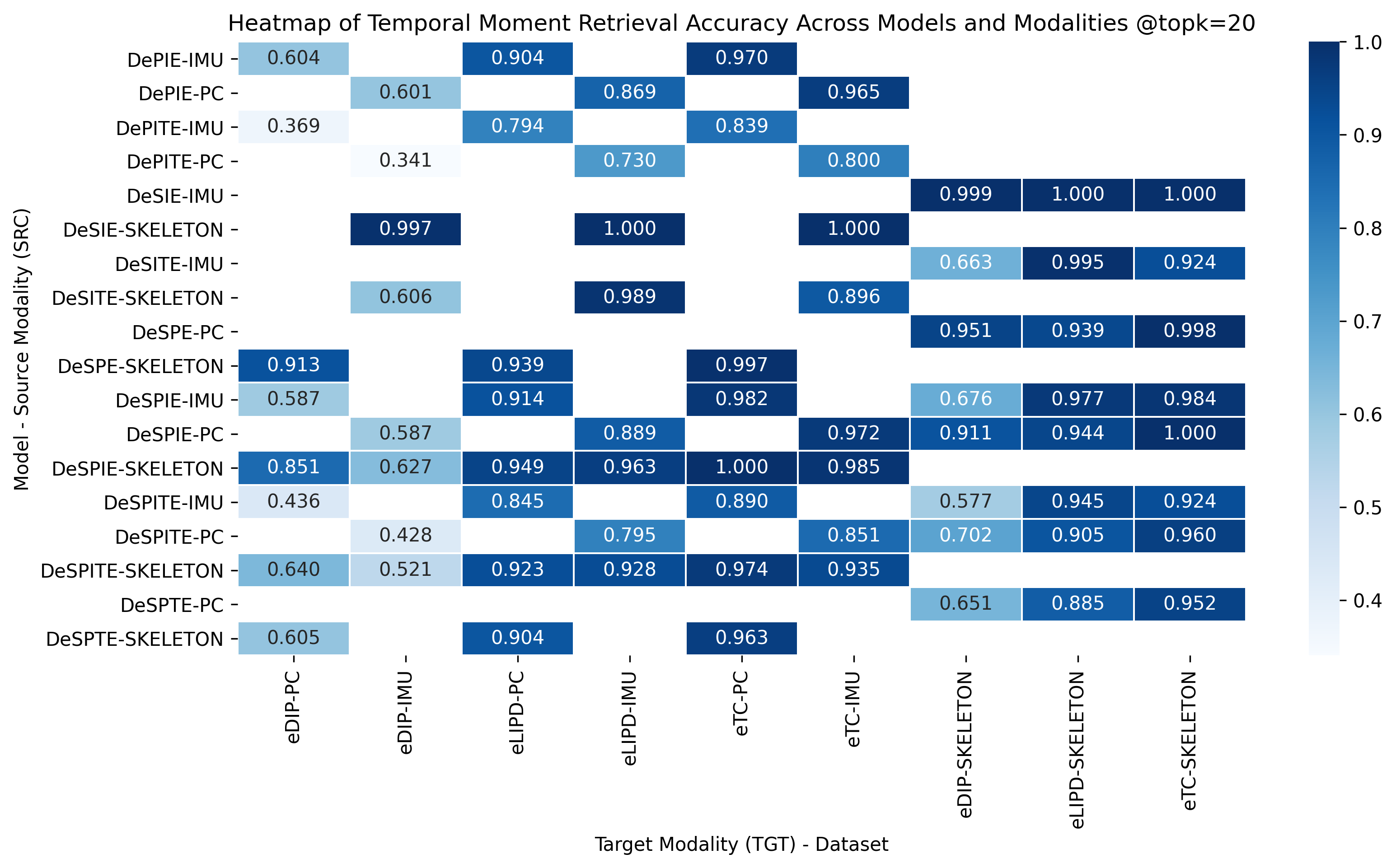}
    \caption{Temporal Moment Retrieval, topk=20}
    \label{fig:enter-label}
\end{figure}

\begin{figure}
    \centering
    \includegraphics[width=\linewidth]{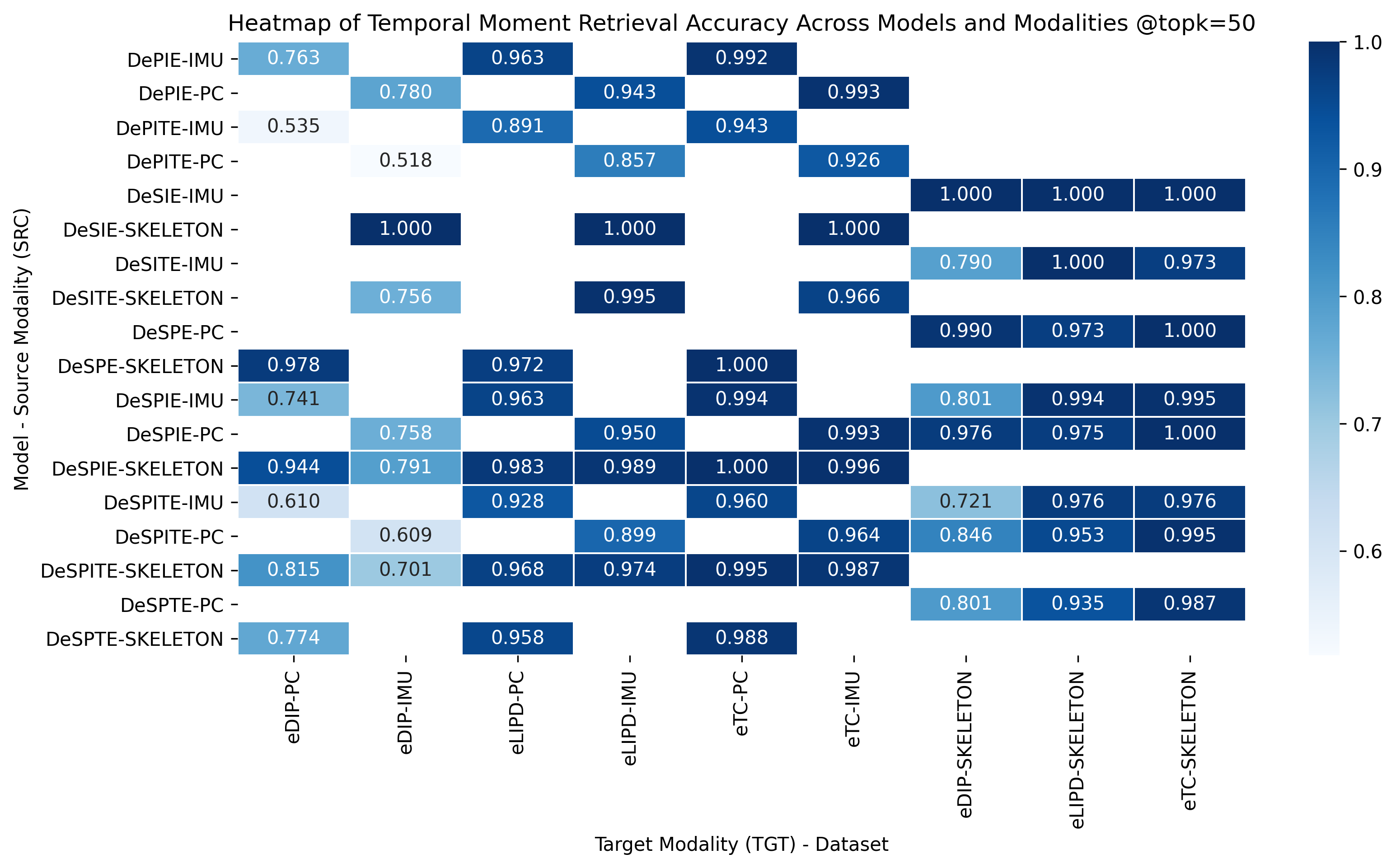}
    \caption{Temporal Moment Retrieval, topk=50}
    \label{fig:enter-label}
\end{figure}

\section{A Simple Improved Matching Algorithm to Associate Different Modalities in the Embedding Space}
In practice, we observe continuous streams of point cloud, skeleton, and IMU time series data. Therefore, a matching score can be computed not only on similarities between a single query sequence and all possible subsequences of a video, but instead on consecutive subsequences. We define such a matching algorithm in Algorithm~\ref{alg:1}, where the mean similarity score over several embeddings from a short temporal neighborhood is considered to compute the temporal matching score. More specifically, given $N$ consecutive subsequences with their respective $Q$ consecutive query embeddings $Z_{a, q}^{j}$, $q \in Q$ from modality $a$ and $K$ consecutive candidate embeddings $Z_{b, k}^{j}$, $k \in K$ for modality $b$, $j \leq 0 < N$. The similarity score between each query $q \in Q$ and candidate $p \in P$ is the average over all pair-wise similarities between the consecutive windows. Finally, the assigned match for each $q \in Q$ is calculated using $argmax$ over all candidate similarity scores. 

\begin{figure*}[t]
    \centering
    \includegraphics[width=\linewidth]{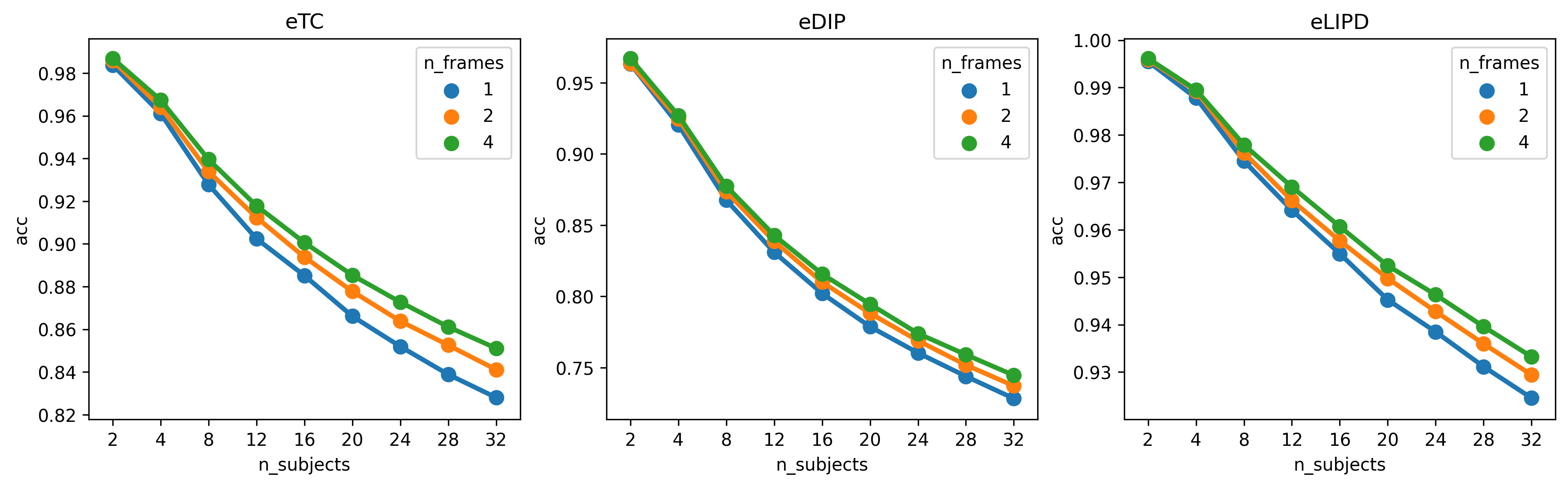}
    \caption{Performance of computing matching scores based on 1,2, or 4 consecutive windows. The matching scores are presented averaged over each model and modality combination to show the effectiveness of matching based on consecutive windows.}
    \label{fig:frame_ablation}
\end{figure*}

Figure~\ref{fig:frame_ablation} presents the average of the results over all modalities for each dataset, respectively. The results verify that observing more frames leads to improved matching results.

\begin{algorithm}[h]
\caption{Matching algorithm for consecutive subsequences}
\label{alg:1}
\KwIn{$Q, K$ consecutive query/candidate embeddings for respective modality $a, b$}
\KwOut{One-to-Many mapping} 

$matches \gets \emptyset$

\ForEach{$q \in Q$}{
    $match \gets \arg\max_{k \in K} \left( \frac{1}{N} \sum_{n=1}^{N} \text{sim}(Z_{a, q}^{j}, Z_{b, k}^{j}) \right)$\;
    $matches \gets matches \cup \{match\}$\;
}
return $matches$
\end{algorithm}

\subsection{Results: Retrieval through Natural Language}
On LIPD-Babel-v2, we evaluate text-to-motion retrieval against TMR++~\cite{bensabath2024cross}. For a fair comparison, we run TMR++ only on the subset of Babel that we use in LIPD-Babel-v2. Note that TMR++ has been trained on the full Babel dataset, which gives the model itself an advantage over our model. 

The results are presented in Table~\ref{tab:text_to_motion}. We use the same threshold (Th) as TMR++ to account for semantically similar retrievals. In addition, we report our results for Th=0.90 and Th=0.95 since the CLIP text embedding space may exhibit different semantic relations compared to the text embeddings used by TMR++. Our results show that our performance is promising, but worse than TMR++, showing possibilities for future work to improve the alignment to text.

\begin{table}[t]
    \centering
    \begin{tabular}{ccc}
        \toprule
        Method & Training Set & R-Top-1 \\
        \midrule
        TMR++ Th=0.95 & Full Babel & 55.54 \\
        \midrule
        Ours (Skeleton) Th=0.95  & \multirow{2}{*}{LIPD-Babel-v2} & 42.55 \\
        Ours (Skeleton) Th=0.90 &   & 48.92 \\
        \midrule
        Ours (IMU) Th=0.95 & \multirow{2}{*}{LIPD-Babel-v2}  & 46.01 \\
        Ours (IMU) Th=0.90 &    & 46.94 \\
        \midrule
        Ours (Pointcloud) Th=0.95 &  \multirow{2}{*}{LIPD-Babel-v2} & 48.68 \\
        Ours (Pointcloud) Th=0.90 &   & 53.62 \\
    \end{tabular}
    \caption{Text-to-$<$Skeleton, Pointcloud, IMU$>$ retrieval on LIPD-Babel-v2}
    \label{tab:text_to_motion}
\end{table}

\section{Ablation Study on LIPD-Babel-v2 for HAR}

We perform a large ablation study between all modalities for downstream classification. We ablate linear/non-linear probing and freezing or fine-tuning each model when training for HAR on LIPD-Babel-v2. The results are presented in Table~\ref{tab:babel_imu} for IMU, Table~\ref{tab:babel_pc} for point clouds, and Table~\ref{tab:babel_skeleton} for skeletons.

%%%% Table should be like this
% Model | Skeleton Point Cloud IMU Text | Probing  Finetuning Projection Head | Acc(seg) 

%%%%%% IMU
\begin{table*}[t]
\centering
\caption{All IMU HAR classification results on the Babel-LIPD-v2-CLS action recognition dataset, segment-level accuracy Acc(Seg) is reported.}
\label{tab:babel_imu}
\begin{tabular}{l|cccc||ccc||c}
\toprule
      model & Skeleton &      PC &     IMU &    Text & Probing & Fine tuning & Projection Head &  $Acc(seg) \uparrow$  \\
\midrule
        Random Init &          &         & $\surd$ &         & $\surd$ &             &     linear &    43.01 \\
Random Init &          &         & $\surd$ &         & $\surd$ &             & non-linear &    57.12 \\
Random Init &          &         & $\surd$ &         &         &     $\surd$ &     linear &    65.62 \\
Random Init &          &         & $\surd$ &         &         &     $\surd$ & non-linear &    64.26 \\
\midrule
\midrule
        PIE &          & $\surd$ & $\surd$ &         & $\surd$ &             &     linear &    60.21 \\
        PIE &          & $\surd$ & $\surd$ &         & $\surd$ &             & non-linear &    62.05 \\
        PIE &          & $\surd$ & $\surd$ &         &         &     $\surd$ &     linear &    68.32 \\
        PIE &          & $\surd$ & $\surd$ &         &         &     $\surd$ & non-linear &    67.50 \\
        SIE &  $\surd$ &         & $\surd$ &         & $\surd$ &             &     linear &    44.20 \\
        SIE &  $\surd$ &         & $\surd$ &         & $\surd$ &             & non-linear &    56.62 \\
        SIE &  $\surd$ &         & $\surd$ &         &         &     $\surd$ &     linear &    66.44 \\
        SIE &  $\surd$ &         & $\surd$ &         &         &     $\surd$ & non-linear &    67.06 \\
        SPIE &  $\surd$ & $\surd$ & $\surd$ &         & $\surd$ &             &     linear &    58.29 \\
       SPIE &  $\surd$ & $\surd$ & $\surd$ &         & $\surd$ &             & non-linear &    60.95 \\
       SPIE &  $\surd$ & $\surd$ & $\surd$ &         &         &     $\surd$ &     linear &    67.28 \\
       SPIE &  $\surd$ & $\surd$ & $\surd$ &         &         &     $\surd$ & non-linear &    69.21 \\
       \midrule
       \midrule
        PITE &          & $\surd$ & $\surd$ & $\surd$ & $\surd$ &             &     linear &    61.14 \\
       PITE &          & $\surd$ & $\surd$ & $\surd$ & $\surd$ &             & non-linear &    59.21 \\
       PITE &          & $\surd$ & $\surd$ & $\surd$ &         &     $\surd$ &     linear &    68.54 \\
       PITE &          & $\surd$ & $\surd$ & $\surd$ &         &     $\surd$ & non-linear &    66.63 \\
       SITE &  $\surd$ &         & $\surd$ & $\surd$ & $\surd$ &             &     linear &    56.69 \\
       SITE &  $\surd$ &         & $\surd$ & $\surd$ & $\surd$ &             & non-linear &    56.95 \\
       SITE &  $\surd$ &         & $\surd$ & $\surd$ &         &     $\surd$ &     linear &    66.86 \\
       SITE &  $\surd$ &         & $\surd$ & $\surd$ &         &     $\surd$ & non-linear &    67.10 \\
      SPITE &  $\surd$ & $\surd$ & $\surd$ & $\surd$ & $\surd$ &             &     linear &    62.06 \\
      SPITE &  $\surd$ & $\surd$ & $\surd$ & $\surd$ & $\surd$ &             & non-linear &    59.76 \\
      SPITE &  $\surd$ & $\surd$ & $\surd$ & $\surd$ &         &     $\surd$ &     linear &    68.08 \\
      SPITE &  $\surd$ & $\surd$ & $\surd$ & $\surd$ &         &     $\surd$ & non-linear &    68.40 \\
\bottomrule
\end{tabular}
\end{table*}

%%%%%% POINTCLOUD
\begin{table*}[t]
\centering
\caption{All point cloud HAR classification results on the Babel-LIPD-v2-CLS action recognition dataset, segment-level accuracy Acc(Seg) is reported.}
\label{tab:babel_pc}
\begin{tabular}{l|cccc||ccc||c}
\toprule

      model & Skeleton &      PC &     IMU &    Text & Probing & Fine tuning & projection &  $Acc(seg) \uparrow$ \\
\midrule
Random Init &          &  $\surd$        & &         &         &     $\surd$ &     linear &    65.69 \\
Random Init &          &   $\surd$      &  &         &         &     $\surd$ & non-linear &    67.38 \\
Random Init &          &   $\surd$      &  &         & $\surd$ &             &     linear &    51.90 \\
Random Init &          &    $\surd$     &  &         & $\surd$ &             & non-linear &    61.98 \\
\midrule
\midrule
        PIE &          & $\surd$ & $\surd$ &         &         &     $\surd$ &     linear &    65.96 \\
        PIE &          & $\surd$ & $\surd$ &         &         &     $\surd$ & non-linear &    69.52 \\
        PIE &          & $\surd$ & $\surd$ &         & $\surd$ &             &     linear &    68.27 \\
        PIE &          & $\surd$ & $\surd$ &         & $\surd$ &             & non-linear &    68.36 \\
        SPE &  $\surd$ & $\surd$ &         &         &         &     $\surd$ &     linear &    66.65 \\
        SPE &  $\surd$ & $\surd$ &         &         &         &     $\surd$ & non-linear &    66.13 \\
        SPE &  $\surd$ & $\surd$ &         &         & $\surd$ &             &     linear &    63.55 \\
        SPE &  $\surd$ & $\surd$ &         &         & $\surd$ &             & non-linear &    64.17 \\
       SPIE &  $\surd$ & $\surd$ & $\surd$ &         &         &     $\surd$ &     linear &    67.51 \\
       SPIE &  $\surd$ & $\surd$ & $\surd$ &         &         &     $\surd$ & non-linear &    66.93 \\
       SPIE &  $\surd$ & $\surd$ & $\surd$ &         & $\surd$ &             &     linear &    67.06 \\
       SPIE &  $\surd$ & $\surd$ & $\surd$ &         & $\surd$ &             & non-linear &    66.41 \\
       \midrule
       \midrule
       SPTE &  $\surd$ & $\surd$ &         & $\surd$ &         &     $\surd$ &     linear &    67.43 \\
       SPTE &  $\surd$ & $\surd$ &         & $\surd$ &         &     $\surd$ & non-linear &    67.30 \\
       SPTE &  $\surd$ & $\surd$ &         & $\surd$ & $\surd$ &             &     linear &    69.31 \\
       SPTE &  $\surd$ & $\surd$ &         & $\surd$ & $\surd$ &             & non-linear &    68.66 \\
       PITE &          & $\surd$ & $\surd$ & $\surd$ &         &     $\surd$ &     linear &    68.84 \\
       PITE &          & $\surd$ & $\surd$ & $\surd$ &         &     $\surd$ & non-linear &    69.50 \\
       PITE &          & $\surd$ & $\surd$ & $\surd$ & $\surd$ &             &     linear &    70.04 \\
       PITE &          & $\surd$ & $\surd$ & $\surd$ & $\surd$ &             & non-linear &    69.00 \\
      SPITE &  $\surd$ & $\surd$ & $\surd$ & $\surd$ &         &     $\surd$ &     linear &    69.00 \\
      SPITE &  $\surd$ & $\surd$ & $\surd$ & $\surd$ &         &     $\surd$ & non-linear &    68.04 \\
      SPITE &  $\surd$ & $\surd$ & $\surd$ & $\surd$ & $\surd$ &             &     linear &    67.06 \\
      SPITE &  $\surd$ & $\surd$ & $\surd$ & $\surd$ & $\surd$ &             & non-linear &    66.32 \\
\bottomrule
\end{tabular}
\end{table*}

%%%%%% SKELETON
\begin{table*}[t]
\centering
\caption{All skeleton HAR classification results on the Babel-LIPD-v2-CLS action recognition dataset, segment-level accuracy Acc(Seg) is reported.}
\label{tab:babel_skeleton}
\begin{tabular}{l||cccc||ccc||c}
\toprule
      model & Skeleton &      PC &     IMU &    Text & Probing & Fine tuning & projection &  $Acc(seg) \uparrow$ \\
\midrule
Random Init &   $\surd$       &         &  &         &         &     $\surd$ &     linear &    67.90 \\
Random Init &   $\surd$       &         &  &         &         &     $\surd$ & non-linear &    68.23 \\
Random Init &    $\surd$      &         &  &         & $\surd$ &             &     linear &    59.71 \\
Random Init &  $\surd$        &         &  &         & $\surd$ &             & non-linear &    60.59 \\
\midrule
\midrule
        SIE &  $\surd$ &         & $\surd$ &         &         &     $\surd$ &     linear &    67.79 \\
        SIE &  $\surd$ &         & $\surd$ &         &         &     $\surd$ & non-linear &    70.44 \\
        SIE &  $\surd$ &         & $\surd$ &         & $\surd$ &             &     linear &    50.50 \\
        SIE &  $\surd$ &         & $\surd$ &         & $\surd$ &             & non-linear &    57.14 \\
        SPE &  $\surd$ & $\surd$ &         &         &         &     $\surd$ &     linear &    69.06 \\
        SPE &  $\surd$ & $\surd$ &         &         &         &     $\surd$ & non-linear &    70.14 \\
        SPE &  $\surd$ & $\surd$ &         &         & $\surd$ &             &     linear &    58.55 \\
        SPE &  $\surd$ & $\surd$ &         &         & $\surd$ &             & non-linear &    59.93 \\
       SPIE &  $\surd$ & $\surd$ & $\surd$ &         &         &     $\surd$ &     linear &    68.31 \\
       SPIE &  $\surd$ & $\surd$ & $\surd$ &         &         &     $\surd$ & non-linear &    67.47 \\
       SPIE &  $\surd$ & $\surd$ & $\surd$ &         & $\surd$ &             &     linear &    61.76 \\
       \midrule
       \midrule
       SPIE &  $\surd$ & $\surd$ & $\surd$ &         & $\surd$ &             & non-linear &    63.64 \\
       SPTE &  $\surd$ & $\surd$ &         & $\surd$ &         &     $\surd$ &     linear &    69.20 \\
       SPTE &  $\surd$ & $\surd$ &         & $\surd$ &         &     $\surd$ & non-linear &    69.01 \\
       SPTE &  $\surd$ & $\surd$ &         & $\surd$ & $\surd$ &             &     linear &    65.84 \\
       SPTE &  $\surd$ & $\surd$ &         & $\surd$ & $\surd$ &             & non-linear &    65.93 \\
       SITE &  $\surd$ &         & $\surd$ & $\surd$ &         &     $\surd$ &     linear &    68.14 \\
       SITE &  $\surd$ &         & $\surd$ & $\surd$ &         &     $\surd$ & non-linear &    69.64 \\
       SITE &  $\surd$ &         & $\surd$ & $\surd$ & $\surd$ &             &     linear &    63.83 \\
       SITE &  $\surd$ &         & $\surd$ & $\surd$ & $\surd$ &             & non-linear &    63.93 \\
      SPITE &  $\surd$ & $\surd$ & $\surd$ & $\surd$ &         &     $\surd$ &     linear &    70.64 \\
      SPITE &  $\surd$ & $\surd$ & $\surd$ & $\surd$ &         &     $\surd$ & non-linear &    69.91 \\
      SPITE &  $\surd$ & $\surd$ & $\surd$ & $\surd$ & $\surd$ &             &     linear &    67.20 \\
      SPITE &  $\surd$ & $\surd$ & $\surd$ & $\surd$ & $\surd$ &             & non-linear &    66.77 \\

\bottomrule
\end{tabular}
\end{table*}

\end{document}